\newif\ifchi
\chifalse 

\ifchi
  \documentclass[manuscript,screen,review,anonymous]{acmart}


\else
  \documentclass[11pt, a4paper, logo, twocolumn, internal, copyright, nonumbering]{googledeepmind}
  \newcommand\Description[2][]{\ignorespaces}
\fi

\usepackage{graphicx}
\usepackage{svg}
\usepackage[most]{tcolorbox}
\usepackage{enumitem}
\usepackage{fontawesome5}
\usepackage{placeins}

\ifchi
\else   
\usepackage[authoryear, sort&compress, round]{natbib}
\fi
\defcitealias{EU_AI_Act_2024}{EU AI Act}


\ifchi
  \title{Human-AI Complementarity: A Goal for Amplified Oversight}

\else
  \title{Human-AI Complementarity: A Goal for Amplified Oversight}
  \correspondingauthor{rishubj@google.com}
  \reportnumber{}
  \renewcommand{\today}{2026-04-23}
  \author[*,1]{Rishub Jain}
  \author[*,1]{Sophie Bridgers}
  \author[$\dagger$,1]{Lili Janzer}
  \author[$\dagger$,1]{Rory Greig}
  \author[$\dagger$,1]{Tian Huey Teh}
  \author[2]{Vladimir Mikulik}
  \affil[*]{Equal first author contributions}
  \affil[$\dagger$]{Equal contributions, order randomized}
  \affil[1]{Google DeepMind}
  \affil[2]{Work done while previously at Google DeepMind}
\fi

\hypersetup{
    colorlinks=true,   
    urlcolor=blue,     
    linkcolor=blue,    
    citecolor=blue     
}

\begin{document}

\begin{abstract}
  Human feedback is critical for aligning AI systems to human values. As AI capabilities improve and AI is used to tackle more challenging tasks, verifying quality and safety becomes increasingly challenging. This paper explores how we can leverage AI to improve the quality of human oversight. We focus on an important safety problem that is already challenging for humans: fact-verification of AI outputs. We find that combining AI ratings and human ratings based on AI rater confidence is better than relying on either alone. Giving humans an AI fact-verification assistant further improves their accuracy, but the type of assistance matters. Displaying AI explanation, confidence, and labels leads to over-reliance, but just showing search results and evidence fosters more appropriate trust. These results have implications for Amplified Oversight---the challenge of combining humans and AI to supervise AI systems even as they surpass human expert performance.
\end{abstract}
\maketitle

\section{Introduction}

Human oversight is critical for ensuring that Artificial Intelligence (AI) models remain safe and aligned to human values. But AI systems are rapidly advancing in capabilities and are being used to complete ever more complex tasks, making it increasingly challenging for humans to verify AI outputs and provide high-quality feedback. How can we ensure that humans can continue to meaningfully evaluate AI performance? An avenue of research to tackle this problem is “\href{https://www.alignmentforum.org/posts/79BPxvSsjzBkiSyTq/agi-safety-and-alignment-at-google-deepmind-a-summary-of#Amplified_Oversight}{Amplified Oversight}”\footnote{\citet{shah2024agi} and \citet{shah2025approach} introduced the term "Amplified Oversight", as the commonly used "Scalable Oversight" may also refer to scaling oversight across a large number of inputs.}, which aims to develop techniques to use AI to amplify humans’ abilities to oversee increasingly powerful AI systems, even if they eventually surpass human capabilities in particular domains.

With this level of advanced AI, we could use AI itself to evaluate other AIs (i.e., AI raters), but this comes with drawbacks (see Section~\ref{sec:oversight}). Importantly, humans and AIs have complementary strengths and weaknesses. We should thus, in principle, be able to leverage these complementary abilities to generate an oversight signal for model training, evaluation, and monitoring that is stronger than what we could get from human raters or AI raters alone. Two promising mechanisms for harnessing human-AI complementarity to improve oversight are:

\begin{itemize}
    \item \textbf{Hybridization}, in which we combine judgments from human raters and AI raters working in isolation based on predictions about their relative rating ability per task instance (e.g., based on confidence), and
    \item \textbf{Rater Assistance}, in which we give human raters access to an AI rating assistant that can critique or point out flaws in an AI output or automate parts of the rating task.
\end{itemize}

In this paper, we empirically investigate how combining Hybridization and Rater Assistance can produce a higher quality oversight signal than relying on either human or AI raters alone. We ground our study in the task of verifying the factual accuracy of AI-generated sentences. We chose this task for two reasons: (1) It is a realistic, pressing safety concern of currently deployed models, which often hallucinate and generate misleading information and (2) It is already a challenging task for human raters \citep{SOPRANO2024103672}.

To do so, we developed an AI fact-verification rater that uses a search engine tool to research factuality. This AI rater has a higher overall accuracy than a typical human rater on our realistic evaluation dataset. But critically, when the AI's confidence is low, it performs worse than humans. This allows us to perform what we call \textit{Confidence-based Hybridization}: use the AI's ratings when it has high confidence, and the humans' ratings when the AI has lower confidence. The AI rater output can also be served to humans as a fact-verification assistant. This approach is a practical implementation of the \textit{selective prediction} or \textit{learning to defer} (L2D) paradigm \citep{geifman2017selective, madras2018predict, mozannar2020consistent}, where a system decides whether to predict autonomously or defer to a human based on reliability.

While both selective prediction and AI-assisted decision making have been studied extensively in isolation, their \textit{interaction}---how the form of AI assistance should change when humans are specifically operating on AI-deferred, low-confidence cases---remains unexplored. Using Rater Assistance to achieve complementarity in settings where AI raters already outperform human raters is challenging; a recent review shows that when AI alone performs better, it is harder for humans to add value \citep{vaccaro2024combinations}. Confidence-based Hybridization can identify a data slice where humans still outperform AI, making it easier for Rater Assistance to achieve complementarity---but the form of assistance that is most effective on the full dataset may not be the most effective on the hybridized slice. We address this gap by systematically studying how 10 different forms of AI assistance affect human accuracy on the specific data slice identified by confidence-based hybridization.

\textbf{This leads to our primary research questions:}
\begin{enumerate}[nosep]
    \item Does confidence-based hybridization improve accuracy above AI or human ratings alone?
    \item On the low-AI-confidence data slice assigned to humans, can AI assistance improve human accuracy, even though the assistant is less accurate than the humans on this slice? If so, what form of AI assistance is most effective at improving accuracy while appropriately calibrating human reliance?
\end{enumerate}

\textbf{Our contributions are:}
\begin{enumerate}[nosep]
    \item We provide the first empirical study of how \textbf{confidence-based hybridization interacts with AI assistance design}. Prior work studies selective prediction \citep{geifman2017selective, bondi2022role} and assistance design \citep{bansal2021does, kim2024imnotsure} as independent problems; we show they are not---hybridization changes which forms of assistance are effective, and the form of assistance most helpful on the full dataset is not most helpful on the AI-deferred slice (Section~\ref{sec:rq2}).
    \item We identify a form of AI assistance (evidence-only, without AI reasoning or judgments) that achieves the rare property of \textbf{helping when the AI is correct without hurting when it is wrong}, breaking the usual over-reliance/under-reliance trade-off \citep{bansal2021does, si2023large}. Where \citet{si2023large} observe that retrieval avoids over-reliance in a general fact-checking setting, we show this property holds \emph{specifically on the low-confidence, AI-deferred slice}---a harder test where the AI's reasoning is more likely to be wrong (Section~\ref{subsec:rq2a}).
    \item We demonstrate that hybridization with evidence-assisted humans achieves 91.3\% accuracy, significantly higher than AI alone (87.7\%), unassisted hybridization (89.3\%), or either party alone---one of few empirical demonstrations of \textbf{complementary team performance} \citep{bansal2021does} in a realistic setting where AI outperforms humans (Section~\ref{sec:t2point5}).
    \item We show that the effectiveness of AI assistance is not static: as raters improve with practice, even evidence-only assistance ceases to help, and leading forms of assistance become actively harmful---a finding not predicted by static learning-to-defer frameworks \citep{madras2018predict, mozannar2020consistent} or prior assistance studies that use single-session designs. This demonstrates that complementarity is a \textbf{moving target} requiring ongoing adaptation (Section~\ref{sec:rq3_summary}).
    \item We ground this study in a \textbf{realistic, challenging AI safety task}---fact-verification of deployed LLM outputs with $N{=}1918$ examples---in contrast to much of the human-AI complementarity literature, which relies on stylized benchmarks (e.g., income prediction, recidivism, image classification, deceptive online review detection) where both the task and the stakes differ substantially from real-world deployment.
    \item We articulate one of the first concrete \textbf{collaboration opportunities between the HCI and AI Alignment communities}: our results show that Amplified Oversight---a core challenge in AI safety---is fundamentally an HCI problem, and that the complementarity methods developed by the HCI community are directly needed to evaluate and align future models.
\end{enumerate}



\section{Related Work}

\subsection{Amplified Oversight}
\label{sec:rw_ao}

As AI capabilities grow, value alignment techniques such as RLHF \citep{christiano2017deepreinforcement} and DPO \citep{rafailov2024direct} depend on humans accurately judging AI outputs across several dimensions  (e.g. helpfulness, factuality, safety). When AI capabilities surpass human ability to evaluate, unreliable feedback risks rewarding behavior that looks superficially "good" but isn't actually desirable \citep{specificationgaming2020}, such as misleading information and sycophancy \citep{sharma2023towards}. Amplified Oversight addresses this by using AI to assist human evaluation~\citep{amodei2016concrete}, through approaches such as AI critics~\citep{mcaleese2024llmcritics}, debate~\citep{irving2018debate, michael2023debate, khan2024debating, kenton2024debate}, and Constitutional AI~\citep{bai2022constitutional, petridis2024constitutionmaker}. Results are promising but mixed: CriticGPT outperformed individual humans but could not help the human-AI team exceed the AI's own performance~\citep{mcaleese2024llmcritics}; debate helps only under knowledge asymmetry~\citep{michael2023debate, parrish2022single, parrish2022two}. We argue that how to use AI to elicit better reasoning from a human rater is fundamentally a Human-Computer Interaction (HCI) problem, and that Amplified Oversight would benefit from closer engagement with HCI research. See Appendix~\ref{sec:extended_rw} for a fuller discussion of proposed techniques and their limitations.

\subsection{Rater Assistance}
\label{sec:rw_ra}

HCI offers extensive research on achieving human-AI complementarity---a term originating in this field. Three themes are especially relevant to Rater Assistance:

\begin{itemize}
    \item \textbf{Synergy is not the default:} Human-AI teams frequently underperform either party acting alone \citep{vaccaro2024combinations}. Complementarity is more likely when humans outperform AI, the AI completes subtasks rather than the entire task \citep{vaccaro2024combinations}, or teams are trained on their respective strengths and weaknesses \citep{pinskiadambenlian2023}.
    \item \textbf{Reliance calibration is a primary obstacle:} \citet{lee2004trust} define \textit{appropriate reliance} as calibrating trust to objective system capability; both overreliance and underreliance degrade team performance \citep{parasuraman1997humans}. Standard interventions---numeric confidence, feature-based explanations---do not robustly improve calibration and can exacerbate over-reliance \citep{bansal2021does, lai2019human, buccinca2020proxy, kaur2020interpreting}. Verbalized uncertainty expressions may be more effective \citep{kim2024imnotsure}, and prompting humans to reflect on their own confidence can help \citep{ma2023should, ma2024}. \citet{schemmer2023appropriate} propose measuring reliance along two dimensions (RAIR and RSR), providing a richer framework for evaluating the over- and under-reliance trade-offs we study. Contrastive explanations reduce over-reliance but increase under-reliance \citep{bansal2021does, si2023large}. In disinformation detection, free-text explanations can improve non-expert accuracy but induce blind trust when the AI is wrong \citep{schmitt2024role}, and multi-step transparent workflows can mitigate blind over-reliance but may introduce under-reliance \citep{he2025finegrained}.
    \item \textbf{Cognitive costs drive behavior:} Over-reliance often reflects a rational cost-benefit trade-off; cognitively forcing analytical engagement---e.g., requiring an initial decision before seeing AI advice---can reduce it \citep{vasconcelos2023explanations, bucinca2021}.
\end{itemize}

\subsection{Hybridization}
\label{sec:rw_hyb}

Hybridization---combining human and AI ratings---can leverage complementarity through two main approaches: (1) averaging or ensembling ratings per datum \citep[see][]{li2024comparative}, and (2) routing tasks to humans or AIs based on predicted reliability, such as AI confidence \citep[see][]{wang2024human}. The latter is studied in ML as \textit{selective prediction} \citep{geifman2017selective} and \textit{learning to defer} (L2D) \citep{madras2018predict, mozannar2020consistent, strong2025learning}. Standard L2D treats humans as static oracles; our two-stage approach \citep{mao2023twostage} challenges this by using Rater Assistance to dynamically improve human competence. Related formalisms include \textit{algorithmic triage} \citep{raghu2019algorithmic, Dvijotham2023} and Bayesian complementarity conditions \citep{steyvers2022bayesian}, which show that decorrelated errors enable hybrid systems to outperform either party alone.

In HCI, confidence-based delegation has improved image classification accuracy over AI alone \citep{hemmer2023, fugener2022cognitive}. However, foundational selective prediction work often ignores that human performance is dynamic and affected by the delegating system itself \citep{bondi2022role, ibrahim2025confidence}. Hybridization has also received less attention from Amplified Oversight than rater assistance, despite the increasing use of AI raters \citep[e.g.,][]{bai2022constitutional}. Crucially, combining Hybridization with Rater Assistance creates an opportunity: confidence-based slicing can identify data where the AI struggles to decide correctly but can still meaningfully assist humans, and the form of assistance most helpful on a particular slice may differ from what works best overall. See Appendix~\ref{sec:extended_rw} for an extended discussion.

\section{Experiments and Results}

\subsection{Overview}

We test these research questions across several waves of experiments on a realistic fact-verification task: rating the factual correctness of AI-generated sentences. We verified factuality at the sentence level rather than whole responses because it was easier to collect high-quality golden (ground-truth) labels.

First, we show that confidence-based hybridization improves overall accuracy above either AI or human ratings alone (RQ1, Section~\ref{sec:hybrid_results}). Next, across 10 between-subjects experiments, we selectively filter the information the AI assistant provides---varying the presence of search results, selected evidence, reasoning, judgments, and confidence---to identify what best calibrates human reliance. We find that less leading assistance (search results and evidence snippets only) avoids over-reliance and is most effective after hybridization, while more leading forms that include factuality labels and confidence scores cause over-reliance (RQ2, Sections~\ref{sec:rq2}--\ref{subsec:rq2a}). Finally, we examine how these effects change as raters improve with practice (RQ3, Section~\ref{sec:rq3_summary}).

These findings can be used directly to improve the quality of factuality evaluations of LLMs, and provide insight for other hard-to-verify, safety-critical evaluation tasks.

\subsection{Overall Methods}

\ifchi
\else
\textbf{Evaluation Set.} The \textit{Evaluation Set} has 1918 [prompt, response, target sentence] tuples in total sampled from a realistic distribution of Gemini responses to User prompts. Each target sentence has a golden label of either "Accurate" or "Inaccurate" from high quality human raters.
\fi

\textbf{Factuality Task.}
Raters assess each target sentence as "Inaccurate", "Unsupported", "Disputed", "Accurate", or "Doesn't require attribution", using online research and/or information from the AI fact-verification assistant (if displayed). Raters can also select "Can't confidently assess", or skip the question, though they are asked to try to avoid doing so. 

\textbf{Analysis.} For all calculations of accuracy, we binarize all ratings into "Accurate" and "Inaccurate," with "Unsupported," "Disputed," and "Doesn't require attribution" considered "Inaccurate."  AI and human ratings are scored as "correct" if they match the golden label and "incorrect" otherwise. We mark ratings with "Can't confidently assess" as incorrect.

\textbf{AI fact-verification model.} We built an AI model that uses a search-engine to research the factuality of a given sentence. This model can serve as both an AI-rater and an assistant to human raters. We trained the AI model to (1) \textit{decompose} the target sentence into its constituent factual claims, (2) \textit{search the web} for reputable evidence to assess the accuracy of each claim, (3) \textit{select quotes} from helpful pieces of evidence, (4) \textit{provide reasoning} about the accuracy of each claim based on the evidence, and (5) \textit{provide ratings} for each claim and the sentence overall. To determine the final Overall AI-Rating, we sample our model 50 times, filter samples by if they pass a programmatic Format Verifier, and take the Majority (modal) Vote of the remaining samples' Overall Ratings. We also determine a model Confidence level on this rating, which is the proportion of verified samples that have the same Overall Rating as the Majority Vote. These confidence scores were fairly well-calibrated (see Fig~\ref{fig:model_calibration}).

\textbf{Human Experiment Set-up.} For all of our human rater experiments, we recruited participants from Prolific, using a pool filtered for general competency but not specifically for fact-verification ability. Experiments were divided into two types: baseline (unassisted) where raters worked independently, and intervention (assisted) where raters were provided information from an AI fact-verification assistant. The T1, T2.5, and T3 experiments all drew from the same rating pool, and within each round all studies were launched simultaneously to reduce timing effects. The T2 experiments used the same pool but split it into exclusive slices so that each sub-experiment had its own set of unique raters with no participant overlap across the 10 T2 conditions.

(See Section \ref{sec:appendix_model_details} in the Appendix for more details on the evaluation set, task, model, and human experiment set-up.)

\subsection{RQ1: Does confidence-based hybridization improve accuracy above AI or human ratings alone?}

Our first research question was whether or not hybridizing between AI ratings and human ratings would lead to higher overall accuracy on the evaluation dataset than using either alone.

We selected which examples to have the human raters rate based on the AI's own confidence. That is, we selected a threshold T, and if the model's confidence was > T on a particular example, then we accepted the AI rating. Otherwise, we had the humans rate it.

\subsubsection{"T1" Human Rater Experiment setup}
\label{sec:t1}

To see if there is a threshold T at which hybridized accuracy is higher than either AI or human accuracy, we ran an unassisted baseline experiment in which human participants rated the entire 1918-example Evaluation Set on their own without assistance (NB: we will refer to this baseline as \textit{baseline\_t1} for time 1). We recruited for 800 rating hours with a maximum of 400 concurrent raters, and each participant could take part for 1 or 2 hours. In total, 475 unique participants took part, each rating 42 examples, for an average of 25.2 ratings per rating-hour. To control for variance in rating speed, we capped used ratings to the first 28 tasks per participant for analysis. Excess ratings were discarded; all reported metrics (e.g., the average of 5.5 ratings per example) apply to this filtered dataset.

\subsubsection{Results}
\label{sec:hybrid_results}

\begin{figure*}[h]
  \centering
  \includegraphics[width=0.7\textwidth]{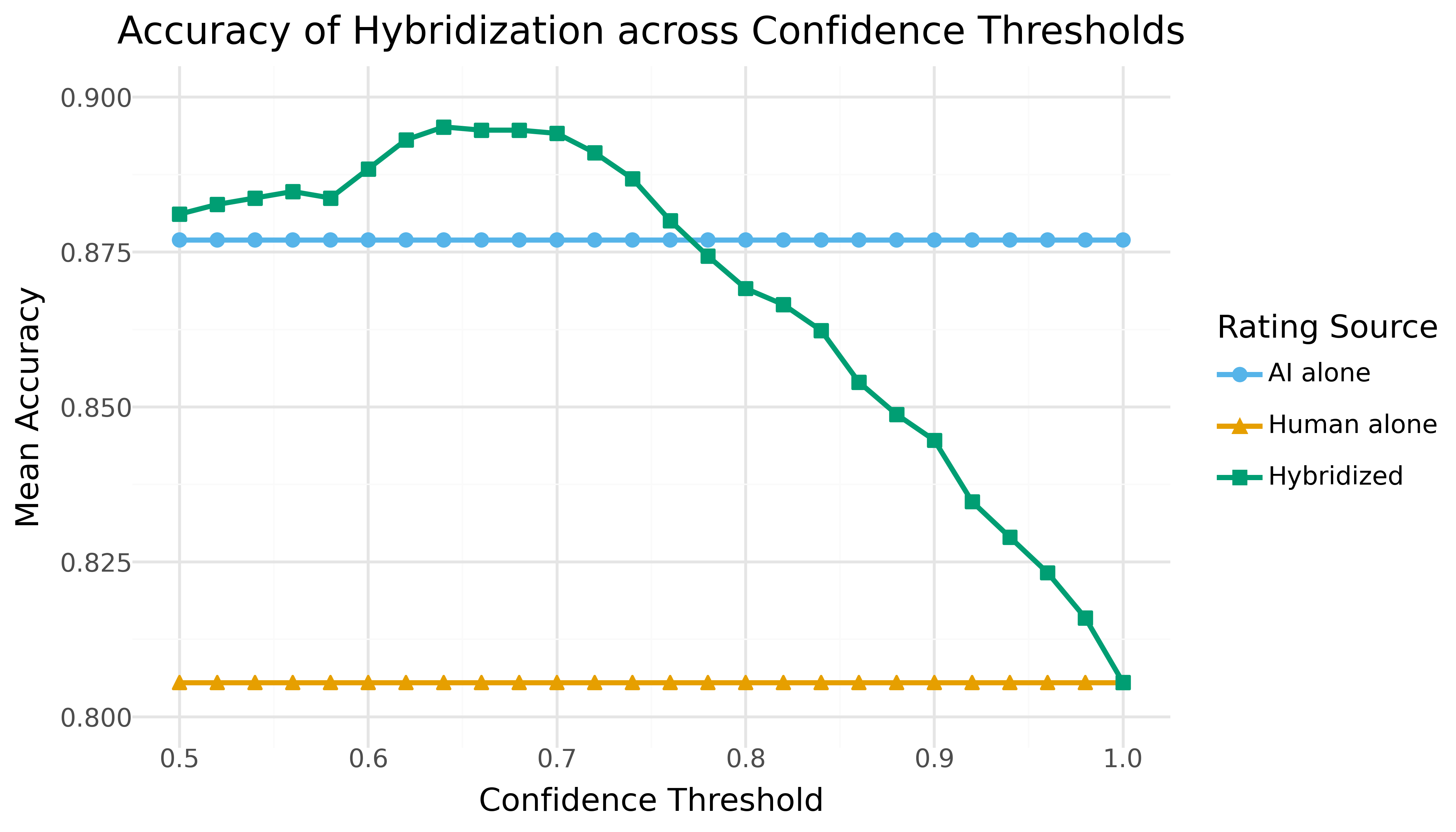}
  \caption{Mean Accuracy of different rating protocols by Confidence Threshold. Humans alone (orange) achieved lower accuracy than AI alone (blue). Hybridized accuracy (green) varies by confidence threshold and there is a range of thresholds for which Hybridization achieves a statistically significantly higher accuracy than using only AI ratings. (NB: Human ratings are majority vote labels from the "T1" studies, described in Section \ref{sec:t1})}
  \label{fig:baseline_hybrid_thresholds}
  \Description{A line chart titled "Accuracy of Hybridization across Confidence Thresholds." The x-axis represents the "Confidence Threshold" from 0.5 to 1.0. The y-axis represents "Mean Accuracy" from 0.800 to 0.900. There are three lines: "AI alone" (blue) is mostly flat at approximately 0.877 until a threshold of 0.75, after which it drops. "Human alone" (orange) is a flat line at approximately 0.806. "Hybridized" (green) starts at 0.87, peaks at 0.893 at a threshold of 0.62, and then declines, crossing below the "AI alone" line at a threshold of 0.74. The chart demonstrates that a hybridized approach can achieve higher accuracy than either AI or humans alone within a specific range of AI confidence thresholds.}
\end{figure*}


On the evaluation dataset, the AI assistant as a rater achieves 87.7\% accuracy, performing above human raters who achieved 75.1\% accuracy on average. Aggregating human ratings into a single label rather than averaging individual ratings has been shown to increase accuracy. Since multiple raters provided ratings for each example, we can calculate the majority vote rating. We take the modal rating as the majority vote rating (for ties, we mark the example as ``Inaccurate''). If we use human majority vote ratings, human raters achieve a higher accuracy of 80.6\% (compared to 75.1\%) on the entire Evaluation Set, but still perform worse than the AI rater. Since human accuracy is higher using majority vote ratings, we will use these aggregated ratings for Hybridization.

If we conduct Confidence-based Hybridization with a threshold T=.62 (i.e., humans rate all examples where AI confidence in the factuality rating is equal to or less than .62), accuracy on the entire Evaluation Set is 89.3\%, higher than using AI ratings alone (87.7\%). As can be seen in Fig.~\ref{fig:baseline_hybrid_thresholds}, there are a range of thresholds below T=0.74 for which Hybridized accuracy is higher than AI alone. For our analyses, we use T=0.62 as this is a threshold among a small set of thresholds where Hybridized accuracy is at its highest.

A mixed effects logistic regression confirmed that while human majority vote was significantly less accurate than AI alone ($p < .001$), Hybridized accuracy was significantly higher than AI ratings alone ($p = .012$). See Appendix~\ref{sec:hybrid_results_details} for full model specifications and coefficients.

Hybridization can only lead to an increase in accuracy compared to AI if humans are better than the AI on the subset of examples they rate. If we filter the Evaluation Set using T=.62, there are 280 examples where AI confidence in the overall rating is 0.62 or lower, what we will call the \textit{Post-Hybridized} Human Set. On this Human Set, AI accuracy is 60.5\% (much lower than accuracy on the entire set), while human accuracy is 71.3\%. To test whether this difference was statistically significantly, we fit a mixed effects logistic regression predicting label accuracy (0 or 1) from a fixed effect of label-type (i.e., AI or human with AI as the reference) and a random intercept by conversation. This analysis indicated that human ratings were indeed significantly more accurate than AI ratings on this set ($\beta = 0.514, SE = 0.185, z = 2.775, p = 0.006$).

\begin{figure*}[t]
  \centering
  \begin{minipage}[b]{0.58\linewidth}
    \centering
    \includegraphics[width=\linewidth]{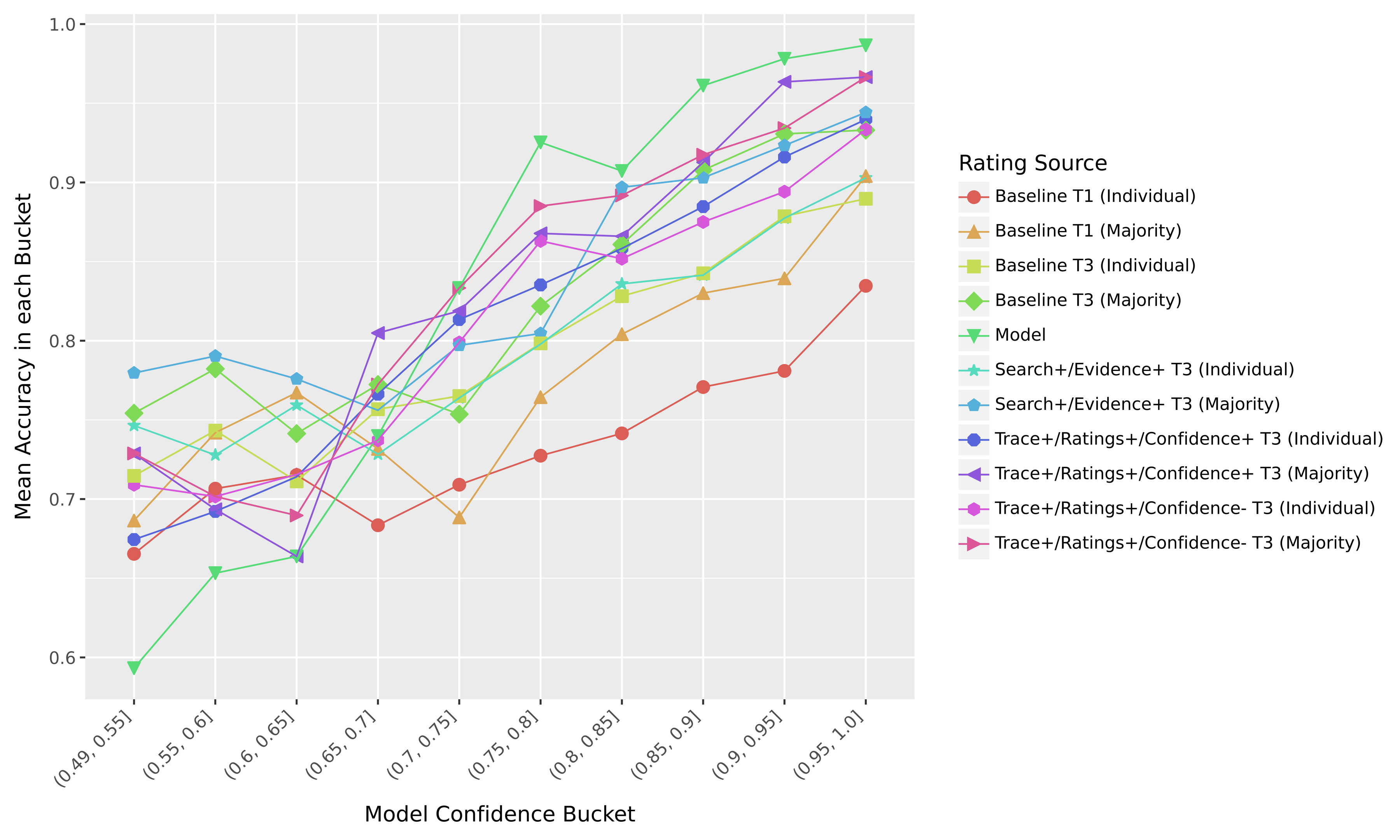}
  \end{minipage}
  \hfill
  \begin{minipage}[b]{0.38\linewidth}
    \centering
    \includegraphics[width=\linewidth]{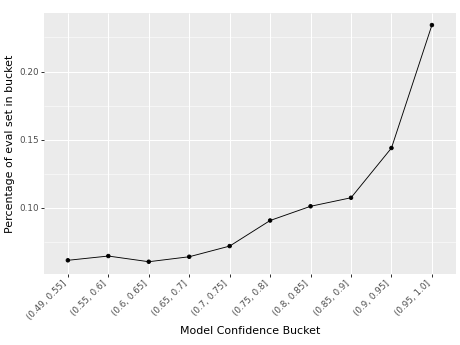}
  \end{minipage}
  \caption{\textbf{Left:} Accuracy bucketed by model confidence. The model is well-calibrated: accuracy increases with confidence. At higher confidences, the model outperforms all forms of human ratings, but at lower confidences, humans are more accurate---enabling complementarity through hybridization. Note: Majority vote accuracies between T1 (Section~\ref{sec:t1}) and T3 (Section~\ref{sec:t3}) are not directly comparable due to differing numbers of ratings per example (5.5 vs.\ 3.61). \textbf{Right:} Distribution of the evaluation dataset across confidence buckets. Most data falls at high confidence, explaining the model's high overall accuracy.}
  \label{fig:model_calibration}
  \label{fig:model_bucket_perc_data}
  \Description{Left panel: A line chart showing accuracy by model confidence bucket. The model line has a strong positive trend, confirming calibration. Human rating lines are flatter. At low confidence, humans outperform the model; at high confidence, the model outperforms humans. Right panel: A chart showing the percentage of evaluation data in each confidence bucket. The distribution is right-skewed, with nearly 25 percent of data in the highest confidence bucket.}
\end{figure*}

In conclusion, confidence-based hybridization enabled human-AI complementarity: combining human and AI labels led to higher overall accuracy on the Evaluation Set than using AI ratings or human ratings alone. This is because (1) we had a fairly well-calibrated confidence metric, and (2) there was some level of decorrelation between human and AI labels (as the AI's accuracy decreased with lower AI-confidence, human accuracy didn't decrease as much), as shown in Fig~\ref{fig:model_calibration}. These two properties may help in identifying which tasks and methods are promising for hybridization.

\subsection{RQ2: Does AI assistance improve human rater accuracy?}
\label{sec:rq2}


Next, we investigated whether providing human raters with access to the AI as a fact-verification assistant would improve their rating accuracy, specifically on the Post-hybridized Human Set. We were also curious what form of assistance would be best for fostering appropriate reliance (Section \ref{subsec:rq2a} below). We ran a series of experiments where human raters were given access to different versions of the AI fact-verification assistant, whose information they could use to help them determine the factuality of the sentence. We had a few different hypotheses about the possible results of these studies. First, it's possible that providing assistance on the hybridised set, the set of examples where the AI is less confident, might actually hurt rather than help raters, especially given that the AI assistant performs statistically significantly worse than human raters on this set. Alternatively, since the fact-verification trace generated by the AI includes search results and verified quotes from websites, it's possible that there is information in the trace that is still helpful to human raters; likewise, if there are errors in the AI's logic or reasoning, this might still be useful to humans for identifying the correct factuality rating. Still another possibility is that humans recognize that they are more competent than the AI and so choose to disregard the information it provides, leading to no difference in assisted vs. unassisted accuracy.

\subsubsection{"T2" Human Rater Experiments}
\label{sec:t2}

We ran 10 experiments, one unassisted baseline experiment (which we will call \textit{baseline\_t2} for time 2) and nine different assisted intervention experiments. This was a between-subjects design, so participants only participated in one of the 10 experiments. We created a pool of Prolific raters and split them into 10 exclusive slices, with each experiment assigned its own slice; no participants overlapped across conditions. Participants were recruited from the same larger pool of raters as the previous experiment, but we took measures to ensure that participants who took part in the previous experiment were evenly distributed across the current experiments.

Unlike the initial baseline experiment, participants did not rate the entire Evaluation Set and instead only rated the examples where model confidence was equal to or lower than .75 (611 examples). We had them rate more than just the 280 examples in the Post-hybridized Human Set (with T=.62), since if assistance improves human performance it's possible that the optimal threshold for hybridization might change.  We ran the baseline experiment again because rater performance may vary depending on time of day, week, etc. Also, rater skill on the given task might change over time, as participants get more practice doing similar tasks.

We recruited for 256 rating hours per experiment with a maximum of 175 concurrent raters, and each participant could take part for 1--3 hours. We collected data for 3 days per experiment, but since each slice only had approximately 80 active raters, we could not achieve the full 256 rating hours requested and received an average of 185 rating hours per study. Across the 10 experiments, there were on average 72 unique participants per experiment, each rating an average of approximately 36 examples, for an average of 14.2 ratings per rating hour spent. To control for rating speed, we capped analyzed ratings to the first 42 tasks per participant, and discarded the rest.

\textbf{Different Versions of the Assistant.} Because we trained the AI fact-verification assistant to output a highly structured output, we could easily filter and curate which aspects of the output were displayed and how. 

We will use the following terminology to describe what information was displayed in each experiment:

\begin{itemize}
    \item \textbf{Search results} - the search queries and results with URLs and snippets.
    \item \textbf{Evidence} - the numbered list of selected evidence with URLs and quotes. Quotes were verbatim from web sources, and usually a couple sentences long. (NB: Whenever Evidence was shown, Search results were also shown)
    \item \textbf{Reasoning} - a list of extracted factual claims from the sentence, and explanation of the factual accuracy of each claim, citing and summarizing the evidence.
    \item \textbf{Judgments} - the judgments of factual accuracy for each claim (if "Reasoning" is included) and the overall rating for the sentence (NB: if factual claims are not listed, then only the Overall Judgment is displayed). 
    \item \textbf{Confidence} - the confidence score expressed as both a range (low, medium, high) and rounded point estimate (e.g., Model Confidence: low (65\%)). Participants were told that they could think of the confidence percentage as the likelihood that the information in the trace led the AI fact-verification assistant to the correct factuality rating for the sentence, which is a reasonable description since the assistant's confidence is fairly well calibrated.
\end{itemize}

For all versions of the assistant, there were drop-down menus under each section that participants could expand to learn more about the information included. They were warned in several places about how the information could be misleading (i.e., "Some claims might be missing, and the summary and evidence could be misleading. Indented Quoted text is guaranteed to be from the webpage. But, this evidence might still be untrustworthy, insufficient, or irrelevant."). If a rating was included, it was described as a predicted rating and raters were warned that it could be incorrect. Again, we hoped such warnings would encourage them to engage more critically with the assistant and not take its reasoning and judgments at face value.

We showed 8 different subsets of the full AI output to the human rater, to understand how these help them at the factuality verification task. See Figure \ref{fig:interventions_indiv_accuracy} for our full list of experiments. We also included an "AI Debate" experiment, where we show the full output (Search, Evidence, Reasoning, and Judgments, but without Confidence), arguing for different overall factuality ratings (i.e., one trace argued for "Accurate" while the other argued for "Inaccurate") - this set-up is analogous to one-turn, simultaneous debate.

\subsubsection{Results}
\label{sec:rq2_results}

\begin{figure*}[h]
  \centering
  \includegraphics[width=\linewidth]{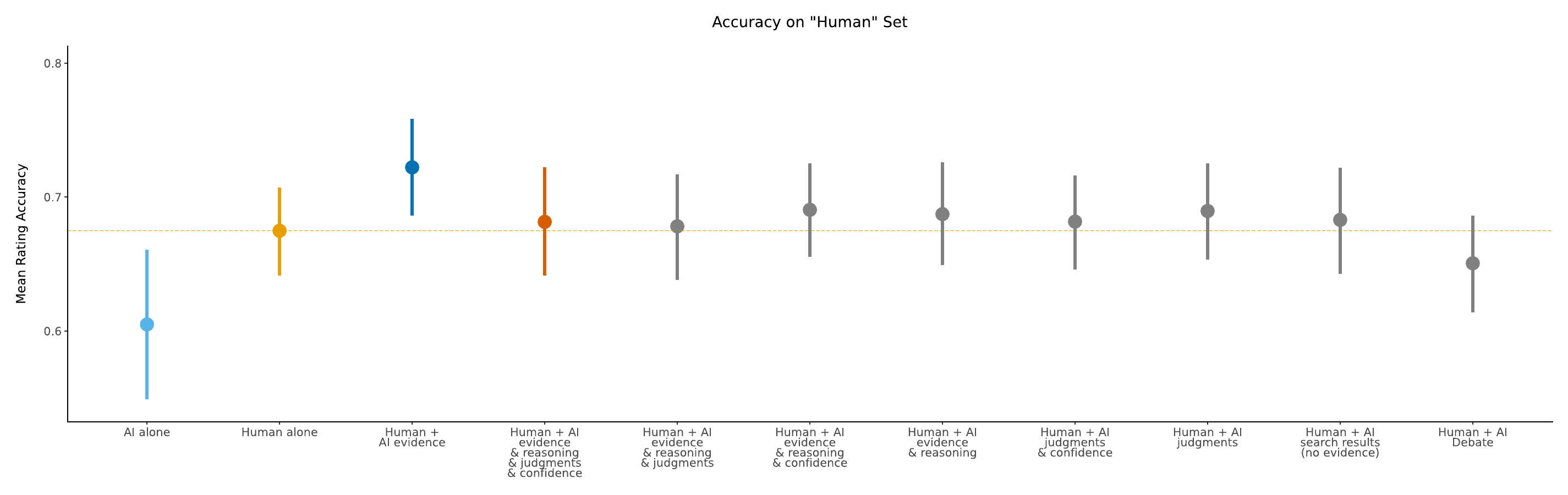}
  \caption{Mean Individual Rater Accuracy, for the different assistant presentations. Accuracy is calculated over the Post-hybridized Human Set (i.e., restricted to examples where model confidence in the overall rating is <= .62). Dotted horizontal lines are Model Accuracy (60\%) and Baseline unassisted accuracy (69.3\%) to ease comparison. Error bars are 95\% Bootstrapped confidence intervals of the mean.  These were from "T2" studies described in Section \ref{sec:t2}.} \label{fig:interventions_indiv_accuracy}
  \Description{A point and error bar chart titled "Accuracy on 'Human' Set." The y-axis shows "Mean Rating Accuracy" from 0.6 to 0.8. The x-axis lists ten different experimental conditions for rater assistance, such as "AI alone," "Human alone," and various "Human + AI" combinations. Dotted horizontal lines indicate the accuracy for "Model Accuracy" (AI alone) at about 60.5\% and "Baseline unassisted accuracy" (Human alone) at 69.3\%. The data point for "Human + AI evidence" is the highest, at 73.3\%, and its 95\% confidence interval error bar does not overlap with the baseline, indicating a statistically significant improvement. Other assistance types are clustered around the baseline, while "Human + AI Debate" is lower.}
\end{figure*}

Fig \ref{fig:interventions_indiv_accuracy} shows how the different assistant presentation styles affect the average human rater accuracy on the Post-Hybridized Human Set (T=.62, 280 examples). We use bootstrap resampling to calculate confidence intervals on the Mean Individual Rater Accuracy for visual benefit. A mixed effects logistic regression on the Post-Hybridized Human Set revealed that most forms of assistance (including Evidence+Reasoning, Judgments, and Confidence) did not yield statistically significant improvements over the baseline (all $p > .22$). The only exception was \textbf{Search and Evidence} assistance, which led to significantly higher average accuracy (73.3\%, $p = .023$),  and performed significantly better than both Judgments \& Confidence ($p = .043$) and Debate ($p = .003$). See Appendix~\ref{sec:rq2_results_details} for full model specifications and coefficients.

All in all these results suggest that for our assistant (1) the selected evidence is particularly helpful for improving overall rating accuracy and (2) the rest of Evidence+Reasoning, Judgments, and Confidence did not appear to statistically significantly or differentially improve rater accuracy. It's possible these components are indeed helpful when they support the correct overall factuality rating, but when overall rating is incorrect, these components may be misleading, limiting the degree to which this information can overall increase rating accuracy. In other words, it's possible that these forms of assistance help raters when the AI is right but hurt raters when the AI is wrong. In the next section, we investigate under- and over-reliance directly by examining assisted rater accuracy for the examples where the AI generated the correct factuality rating for the sentence vs. examples where the overall rating was incorrect.

\subsubsection{RQ2a: What form of assistance is best for calibrating reliance?}\label{subsec:rq2a}

To understand why certain forms of rater assistance work better than others, it is useful to compare the degree to which human raters under- and/or over-rely on the assistant. That is, does the rater use or trust the information the assistant provides the appropriate amount, too little, or too much given its validity? We operationalize under-reliance, i.e. not trusting the assistant enough when it is warranted to do so, as how often the rater still provides the incorrect factuality judgment for the sentence when the assistant generates the correct label. Conversely, we operationalize over-reliance, i.e. trusting the assistant too much when it's not warranted, as how often the rater provides the incorrect factuality judgment when the assistant generates the incorrect label compared to how often the rater provides the incorrect judgment without assistance (i.e., does the assistant significantly decrease raters' performance on this slice of the data).

As identified in the HCI literature, over-reliance is particularly important to avoid, as well as especially critical for achieving complementary performance. One would hope that the assistant only provides additional information, and doesn't persuade the rater to incorrectly go against their own judgment or to perform worse than they would on their own, ultimately, reducing the benefits of hybridization as raters are less likely to rely on their own complementary strengths. In a future where rater assistants are widely deployed, we don't want raters to uncritically and across-the-board default to the AI's output. We want to leverage the complementary skills and knowledge of AI and humans to achieve a higher quality supervision signal than either entity could achieve alone. Ideally, we would like a form of assistant that helps when it is correct and at least does not hurt when it is incorrect.

\begin{figure*}[h]
  \centering
  \includegraphics[width=\linewidth]{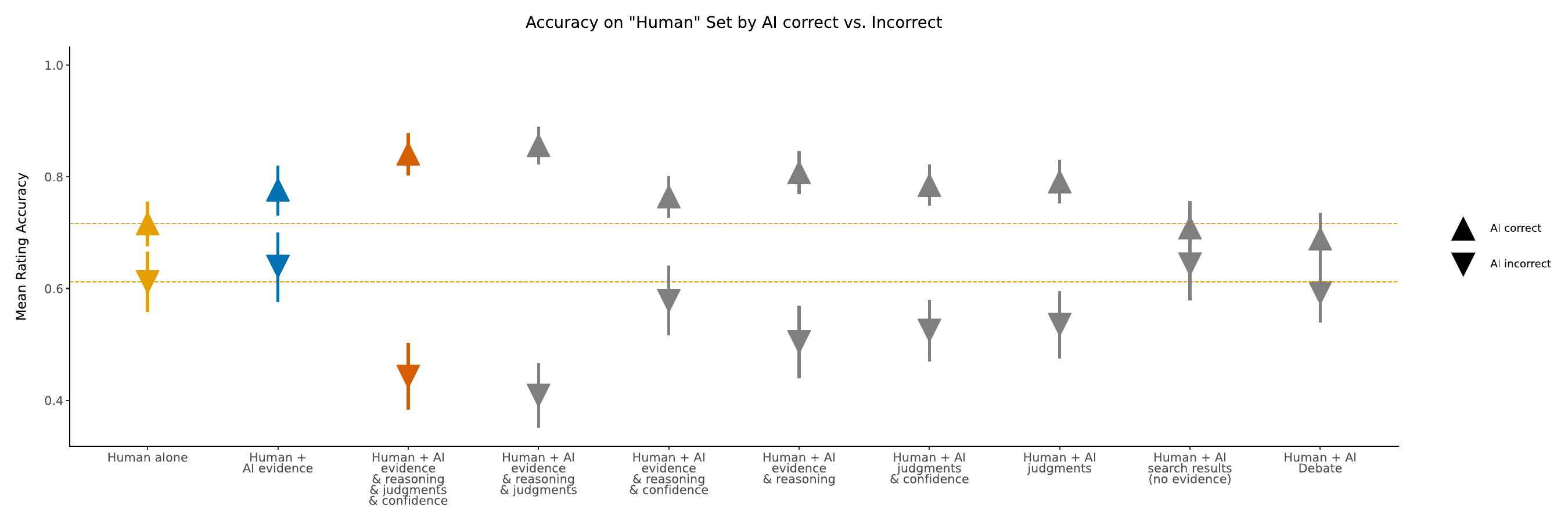}
  \caption{Mean Individual Rater Accuracy, for the different assistant presentations, split by whether the Overall Judgment generated by the fact-verification assistant was correct or incorrect (note that some forms of assistance did not display this judgment). Accuracy is calculated over the Post-hybridized Human Set (i.e., restricted to examples where model confidence in the overall judgment is <= .62). Dotted lines are Baseline accuracy for correct and incorrect to ease comparison. Error bars are 95\% Bootstrapped CIs of the mean. These were from "T2" studies, described in Section \ref{sec:t2}.}
  \label{fig:overtrust_interventions}
  \Description{A grouped point and error bar chart titled "Accuracy on 'Human' Set by AI correct vs. Incorrect." The y-axis is "Mean Rating Accuracy" from 0.4 to 1.0. The x-axis shows the same experimental conditions as Figure 2. For each condition, two data points are plotted: one for when the AI assistant's judgment was correct (blue) and one for when it was incorrect (orange). Dotted lines show the baseline accuracy for correct and incorrect cases. The chart illustrates over-reliance when an orange point is significantly below the baseline (e.g., for "Human + AI evidence \& reasoning \& judgments"), and it illustrates effective assistance when a blue point is significantly above the baseline without a corresponding drop for the orange point (e.g., for "Human + AI evidence").}
\end{figure*}

Figure \ref{fig:overtrust_interventions} breaks down accuracy across the different experiments by whether or not the AI assistant generated the correct overall factuality judgment for the sentence. We analyzed over-reliance using a mixed effects logistic regression. This analysis indicated that when the model Judgments and Evidence+Reasoning were shown (regardless of whether or not Confidence was included), raters performed worse with assistance compared to baseline on examples where the model was incorrect, and this decrement was statistically significant ($p < .001$), i.e., there was evidence of over-reliance. This was also true when both the overall Judgment and Confidence were shown, even if the Evidence and Reasoning are not included ($p = .042$). However, if just the overall Judgment is shown alone, or if no Judgments are included, assisted performance was numerically but not statistically worse than baseline (all $p > .054$). All together these results suggest that if the overall Judgment for the sentence is shown along with another piece of information from the assistant model (Evidence+Reasoning or Confidence), human raters tend to over-rely on the model's judgment. See Appendix~\ref{sec:rq2a_details} for full model specifications and coefficients.

If we turn to our less leading forms of intervention, we see that for Debate, Evidence, and Search-only (not including evidence), performance on the examples that the model gets incorrect is no different when assisted vs. unassisted, i.e., there is no evidence of over-reliance in these experiments (all $\beta$'s between $-0.138$ and $0.163$, and all $p's > .181$). Indeed, for Evidence and Search results (without evidence) performance is even numerically higher than baseline (though the increase in each case is small and not statistically significant). These findings confirm that these forms of assistance are indeed less leading and so are also less mis-leading. Interestingly, showing the full output (without confidence) leads to over-reliance, but showing two of these outputs and contrasting Judgments in the Debate experiment seems to mitigate this tendency.

To better understand the full impact of the fact-verification assistant, it is important to also understand its effect on examples where it generated the correct overall Judgment. If we re-fit the same regression model described above, but refactored so that "correct" is the reference level for the predictor of model accuracy, we find that assisted performance is statistically significantly better than baseline for all forms of assistance (all $\beta$'s $> 0.357$ and all $p's < 0.027$, except for Debate and only showing search results (Debate: $\beta = -0.115, SE = 0.167, z = -0.689, p = .491$; Search-results-only: $\beta = 0.25, SE = 0.172, z = 0.065, p = .948$). 

No form of assistance gets raters to 100\% on the examples where the model generates the correct factuality judgment, so there is evidence of under-reliance throughout, but it appears to be the most extreme for Debate and Search-results-only where raters are no different from baseline - indeed for these forms of assistance they neither help when correct nor hurt when incorrect, so it is possible raters just ignored them. Under-reliance appears to be the least prominent when raters see the Reasoning in combination with the Judgments (regardless of whether Confidence is included) - raters achieve 84\% accuracy when Confidence is included (Evidence \& Reasoning \& Judgments \& Confidence) and 86\% when not (Evidence \& Reasoning \& Judgments). These forms of assistance, however, also exhibit the worst amount of over-reliance, suggesting that they are both the most helpful when correct and the most harmful when wrong, which is why they don't lead to an overall increase in accuracy above baseline.

The only form of assistance that achieves the ideal of helping when correct and not hurting when wrong, is showing just the Evidence (alongside the Search results). This form of assistance statistically significantly improves performance compared to baseline on the examples that the model gets correct (79.3\% vs. 71.3\%, $\beta = 0.446, SE = 0.170, z = 2.631, p = .009$), and as described above, does not affect performance when the model is incorrect (64.0\% vs. 61.5\%). It's possible that when the assistant generates the incorrect factuality judgment, the information present in the search results and evidence list is still useful to the raters and perhaps even correct and valid. If this is the case, it points to a limitation in the way we operationalize over-reliance. We slice the data based on if the final model judgment for the sentence is correct, not based on whether all of the intermediate steps and information provided is correct. Doing the latter would provide a more accurate measure of over-reliance. We return to this limitation in the discussion.

\subsubsection{RQ2 Summary}
\label{sec:t2point5}

In summary, most AI-assistance presentations did not improve accuracy on the Post-Hybridized Human Set, but showing participants the AI's search results and selected evidence \textit{did} increase human rating accuracy because it was best at calibrating reliance. 

For RQ1, Confidence-based Hybridization with unassisted majority vote human ratings achieved higher performance than using AI ratings alone. Unlike baseline\_t1 data, in the experiments for RQ2 (T2 data), we did not collect a sufficient number of ratings per example to see an increase in performance from aggregating individual ratings into a majority label, so we ran an additional intervention experiment using AI Evidence assistance to increase the number of ratings per example. 

The experiment (``T2.5'') was identical in structure and content to the prior search and evidence experiment except that we made it available to the entire pool of participants so that we could achieve sufficient replication per example (i.e., all participants from prior experiments were eligible to participate again). We recruited for 375 rating hours with a maximum of 187 concurrent raters, and each participant could take part for 1 or 2 hours. In total, 264 unique participants took part, each rating an average of approximately 21 examples, for an average of 14.8 ratings per rating hour spent; for analysis, we capped used ratings to the first 28 tasks per participant and discarded the rest. Participants rated examples in the equal to or lower than .75 confidence set, and average replication was 6.42 per example in the analyzed dataset. To ensure a fair comparison of majority voting between T1 and T2.5, we clamped the maximum number of ratings per example to the minimum across both studies. Average individual label accuracy on the Post-Hybridized Human Set (T = 0.62) was 75.12\% and average majority label accuracy was 84.51\%, so we again see a boost from aggregating human labels, as we did for RQ1.

Hybridizing with majority vote Human + AI Evidence ratings from this experiment, using a confidence threshold of 0.62, results in an accuracy of 91.3\%, compared to 89.3\% using unassisted, baseline\_t1 ratings. To test the statistical significance of this difference, we conducted a mixed effects logistic regression on the entire evaluation set predicting label accuracy (1 or 0) from a fixed effect of rating protocol (four-level factor: AI alone, human alone, hybridized w/ unassisted humans, hybridized w/ evidence-assisted humans, with hybridized w/ unassisted humans as the reference) with random intercepts by example and by label source (i.e., AI, unassisted human, assisted human). This analysis indicated that hybridizing with evidence-assisted humans achieved significantly higher accuracy on the entire Evaluation Set than hybridizing with unassisted humans ($\beta = 0.590, SE = 0.180, z = 3.276, p = .001$).

\begin{figure*}[h]
  \centering
  \includegraphics[width=0.4\linewidth]{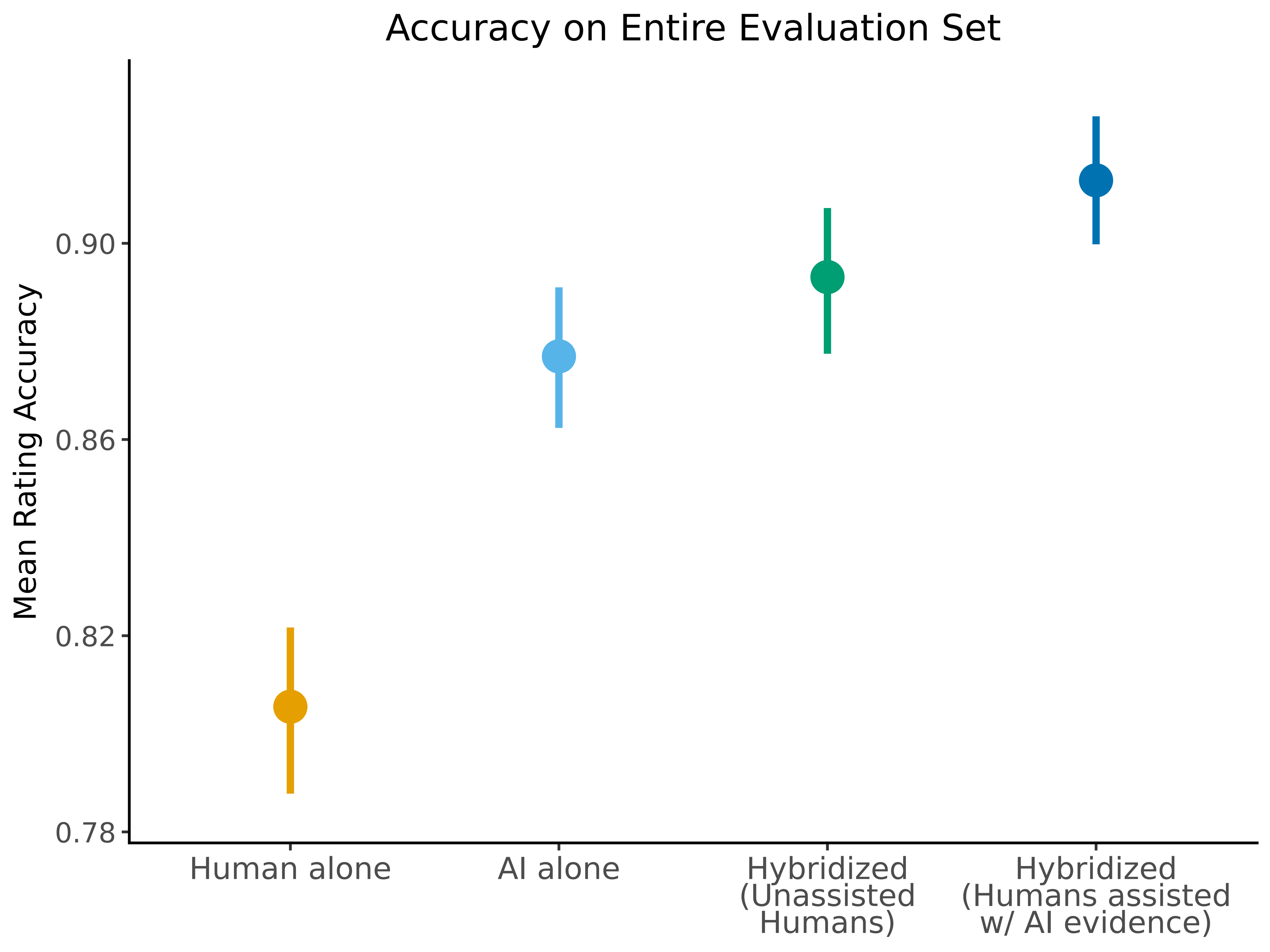}
  \caption{Mean accuracy and 95\% confidence intervals on the entire evaluation set of humans alone (yellow), AI alone (light blue), hybridization with unassisted humans (green), and hybridization with Evidence-assisted humans (dark blue). The assisted-hybridized approach is significantly more accurate than the Human alone, AI alone, and the unassisted-hybridized protocol.}
  \label{fig:all_entire_acc}
  \Description{A point and error bar chart titled "Accuracy on Entire Evaluation Set." The y-axis is "Mean Rating Accuracy" from 0.78 to 0.90. The x-axis shows four approaches: "Human alone" (yellow, ~81\%), "AI alone" (light blue, ~88\%), "Hybridized (Unassisted Humans)" (green, ~89\%), and "Hybridized (Humans assisted w/ AI evidence)" (dark blue, ~91\%). The error bars indicate that each successive approach provides a statistically significant improvement in accuracy over the previous one.}
\end{figure*}

We thus find that Confidence-based Hybridization achieves human-AI complementarity and that combining hybridization with Rater Assistance further increases performance above AI ratings alone. See Figure \ref{fig:all_entire_acc}.

\FloatBarrier

\subsection{RQ3: Does the effectiveness of assistance change as raters improve?}
\label{sec:rq3_summary}

As described in Contribution~4, a key question for the long-term viability of human-AI complementarity is whether it persists as human raters gain experience. To test this, we ran a third wave of experiments (``T3'') with the same rater pool, now more practiced at fact-verification (see Section~\ref{sec:RQ3} for full details). We recruited for 800 rating hours per experiment with a maximum of 400 concurrent raters, and each participant could take part for 1 or 2 hours. Across the 4 T3 experiments, there were on average 471 unique participants per experiment, each rating an average of 28 tasks, for an average of 13.0 ratings per rating hour spent. For analysis, we capped used ratings to the first 28 tasks per participant and discarded the rest.

Human baseline accuracy improved substantially between waves: unassisted raters achieved 82.7\% on the Evaluation Set (individual labels) in T3, compared to 75.1\% in T1. Confidence-based hybridization continued to improve accuracy above either party alone, reaching 90.3\% (majority vote at T=.62). However, AI assistance no longer provided an additional benefit. Evidence-only assistance (\textbf{Search+}/\textbf{Evidence+}), our best-performing intervention in RQ2, produced no statistically significant improvement over the unassisted baseline on the Post-Hybridized Human Set ($p = .647$). More strikingly, leading forms of assistance that included the model's judgments and confidence became \textit{actively harmful}: \textbf{Trace+}/\textbf{Ratings+}/\textbf{Confidence+} significantly decreased accuracy below baseline ($\beta = -0.317, p = .004$).

These results demonstrate that human-AI complementarity is a \textbf{moving target}. As raters improve, the assistance that was once helpful becomes unnecessary or counterproductive, suggesting that assistance design must be continuously adapted to the evolving skill level of the human raters.

\section{Discussion}

\subsection{The Future of Human Oversight in LLM Training and Evaluation}
\label{sec:oversight}
As AI continues to improve, including in rating ability, a critical question arises: will humans even be necessary for oversight in the future? If AI raters become demonstrably superior, will there be any slice of data where assisted humans add value? There are several reasons why we expect it to remain critical to keep humans in-the-loop to develop superhuman AI safely and in line with human values:

\begin{itemize}
    \item \textbf{Capabilities:} The relative strengths of humans compared to AI might change over time, but it is possible that humans will still retain complementary knowledge, skills, and abilities. For example, AIs might still have specific weaknesses such that the human-AI oversight team is more adversarially robust to reward hacking targeting the weaknesses of one or the other specifically (e.g., the comprehensiveness-hallucination trade-off described in \citealp{mcaleese2024llmcritics}). Humans might also be more adversarially robust in general. The current frontier of AI is also very "jagged", and while AI might be better than humans overall for a domain, it might unexpectedly do worse on certain problems that we can identify via tools like AI-confidence.
    \item \textbf{Value Alignment:} Ultimately, humans will likely need to continue to provide input to confirm that AI systems are indeed acting in accordance with human values. This is because human values continue to evolve. In fact, human preferences define a “slice” of data where humans are definitionally more accurate than non-humans (including AI). AI systems might get quite good at predicting what aligned behavior should be in out-of-distribution scenarios, but it’s unlikely that AI will be able to figure out what humans want in completely new situations without humans being consulted and kept in the loop.
    \item \textbf{Trust:} As AIs improve, they might develop the capability to “scheme” and sabotage the rating process \citep{meinke2024frontier}. AIs with these capabilities pose loss of control risks \citep{ngo2022alignment}, and so we should not fully trust the output from these models (including AI raters) without better ways for humans to monitor and supervise them. Humans are capable of scheming and sabotage as well, but we have a lot more experience with endowing trust to humans as part of critical decision making systems and guarding these systems from bad human actors.
\end{itemize}

How humans remain in the loop might change over time. Even if we switch entirely to AI raters for training our models, we may still use human input to train the AI raters and human judgments to evaluate them (as is the case today). If we switch to even higher level approaches such as Constitutional AI \citep{bai2022constitutional}, human input will still be needed to write the rules for the AI to follow. Complementarity will continue to be relevant as long as humans are involved, wherever and however that might be (e.g., AI assistance for constitution writing), and so Amplified Oversight research will continue to benefit from research in HCI and other related disciplines. But the constant improvement of model capabilities will render human-AI complementarity a moving target, which means we need to develop repeatable and generalizable processes for conducting this type of research.


\subsection{Discussion, Limitations, and Future Work}

Overall, our rater AI-assistance with confidence-based hybridization leads to complementary performance, better than either humans or AI individually, in our Factuality rating task where AI-raters outperform human raters.

Section \ref{sec:rq3_summary} shows that at a confidence threshold of 1 (i.e. no confidence-based hybridization, only human ratings), AI-assisted humans perform better than unassisted humans, but they do not surpass model performance. In our experiments, confidence-based hybridization techniques are critical to achieve complementarity, and also critical to enable rater assistance to achieve complementarity. The reason it is hard for even assisted humans to surpass our AI rater seems largely due to the fact that the model gets extremely high accuracy at high confidences (see Figure~\ref{fig:model_calibration}), which it has for a large proportion of the data (see Figure~\ref{fig:model_bucket_perc_data}). Even so, the errors of the AI and humans are not entirely correlated and there is a slice of data where humans do perform better, suggesting the AI and humans have complementary skills on this task that we were able to leverage via confidence-based hybridization.

Importantly, confidence-based hybridization requires a well calibrated AI rater, where increased confidence translates to increased rating accuracy and quality. Proper estimation and calibration of LLMs' uncertainty is a challenging task, and may not be a given for many AI raters. But on tasks where AI raters outperform assisted humans, any sort of slicing of the data to find sections where assisted humans outperform AI raters would lead to increased performance. Exploring other forms of hybridization could be a promising area for future Scalable Oversight research. A key limitation of our confidence-based hybridization approach was that we didn't use a separate "train" set to calibrate the threshold. Future work should understand how well this hybridization method generalizes.

Figure~\ref{fig:model_calibration} shows that some types of assistance do better than AI Evidence at higher confidences, indicating that perhaps hybridization between different forms of assistance might lead to even better performance. In our experiments (see Appendix Section \ref{sec:t3}), this "assistance-hybridization" did not improve accuracy above hybridizing with a single form of assistance, likely because the model still surpasses the other forms of assistance at higher confidence. But we think this "assistance-hybridization" is still a promising research direction, allowing for different types of assistance targeting different types of data.

Our results on debate showed that it led to low over-reliance, yet also low overall accuracy. On the one hand, this might point to debate as a promising method for reducing over-reliance. On the other hand, it didn't lead to \textit{appropriate} reliance as it increased under-reliance, and didn't increase accuracy over the baseline. Our debate method could be improved significantly though. Our model was not trained to give arguments for both sides. Also, there was a lot of text shown to raters (double that of the next longest form of assistance), which could have easily led to raters disengaging and ignoring the assistance.

A critical insight from our experiments is that \textbf{hybridization and assistance are not independent interventions}. On the post-hybridized data slice---where the AI's confidence is low and it performs significantly worse than humans (60.5\% vs.\ 71.3\%)---the AI's reasoning and judgments are more likely to be wrong, making leading forms of assistance actively harmful. This explains why Search+Evidence, which strips the AI's conclusions and preserves only its information-gathering work, is uniquely effective post-hybridization. This finding has practical implications: researchers and practitioners designing human-AI oversight systems should not evaluate assistance designs in isolation, but rather in the context of the specific data routing that precedes them. Unlike \citet{bondi2022role}, who study the effect of deferral messaging on human accuracy, we vary the \textit{form} of post-deferral assistance across 10 conditions and show that this design choice is the primary lever for achieving complementarity.

More broadly, the divergent reliance profiles across our assistance conditions suggest that the \textit{form} of AI assistance---not merely its accuracy---shapes human reliance. This aligns with Miller's (\citeyear{miller2023explainable}) \textit{Evaluative AI} framework and \citet{bucinca2021}'s cognitive forcing functions, demonstrating that structuring assistance as retrievable evidence—rather than passive recommendations—successfully forces analytical engagement. Where prior work has studied verbalized uncertainty as a lever for reliance calibration \citep{kim2024imnotsure}, or contrasting explanations for fact-checking \citep{si2023large}, we find that the \textit{structural form} of evidence---presenting search results and quotes without AI conclusions---achieves the rare ideal of helping when the AI is correct without hurting when it is wrong. This breaks the over-reliance/under-reliance trade-off that \citet{bansal2021does} and others have documented, and provides an empirical demonstration of complementary team performance \citep{steyvers2022bayesian} in a domain where AI outperforms unassisted humans.

Furthermore, our longitudinal findings (Section~\ref{sec:rq3_summary}) reveal that the effectiveness of AI assistance is not static---as human raters improve through practice, even evidence-only assistance ceases to provide a significant benefit, and leading forms of assistance become actively harmful. Unlike L2D frameworks \citep{madras2018predict, mozannar2020consistent} that treat the human as a static oracle, our results show that complementarity is a \textit{moving target} that requires ongoing adaptation of both the hybridization threshold and the assistance design.

As AI improves, the quality of human raters will also need to improve to be able to give useful feedback. Our results around bonus studies (Section \ref{sec:bonus_desc} in Appendix) allude to how the rater assistant we built may not be useful for higher quality raters. In addition, in our RQ3 experiments (Section~\ref{sec:rq3_summary}), raters got better and assistance no longer helped. These practice effects and human raters improving over time could have been another reason that our human rater accuracy in Section \ref{sec:t2point5} was so high, as we opened up that experiment to all prior participants in order to get sufficient number of ratings. Different kinds of assistance might be necessary as raters' abilities on a task improve either from increased motivation or skill. For example, future work could explore building AI assistance that aims to teach human raters, and improve rater quality over time. This could build on the wealth of existing pedagogy research.

The AI assistants will also improve as the capabilities of the underlying foundation models improve. Unfortunately, it will be increasingly difficult to conduct realistic hybridization and rater assistant research, since it will become harder to collect golden data from experts to evaluate how our techniques are improving rater quality. There might already be domains where currently we don't have a good way of getting golden labels (e.g. on moral reasoning questions). If AI becomes so capable that we cannot distinguish the quality of their ratings from experts', new methods of collecting golden labels might need to be created in order to continue to know how best to leverage complementary human and AI strengths.


In addition to Amplified Oversight, Human-AI Complementarity is also a \textbf{requirement} for effective Human Oversight \citepalias[Article 14]{EU_AI_Act_2024}, preventing disempowerment \citep{kulveit2025gradualdisempowermentsystemicexistential}, increasing human agency \citep{jha2026human}, and delaying mass job-loss \citep{acemoglu2020wrong}. Without complementarity, if AI alone consistently outperforms human-AI teams (the default trajectory), there is little rationale for retaining humans in most decision-making loops. Complementarity research thus stands as one of the few technical levers we have to address this seemingly policy-only problem.


\ifchi
\else

\subsection{Acknowledgements}

We would like to thank Geoffrey Irving and Jonathan Uesato for their thought and research leadership on the GDM Rater Assist team; we additionally thank Jonathan for his engineering contributions, which were the foundations for the ongoing work presented here. We are also grateful to John Aslanides and Zoe Ashwood for their crucial contributions to an earlier iteration of the project, and to Rohin Shah, Noah Goodman, Anca Dragan, and Sasha Goldshtein for their thoughtful guidance. Lastly, a big thank you to MH Tessler, Mevan Babakar, Verena Rieser, Irene Rae, Tom Everitt, Noah Siegel, Taylan Cemgil, Zachary Kenton, and Samuel Albanie for their helpful feedback and discussions.

\fi

\pagebreak
\clearpage
\bibliography{main}

\pagebreak
\clearpage
\appendix 
\section{Appendix}

\subsection{Extended Related Work}
\label{sec:extended_rw}

\subsubsection{Amplified Oversight: Additional Techniques}

Beyond the approaches discussed in the main text, several Amplified Oversight techniques have been proposed but not yet evaluated on realistic tasks, including Iterative Amplification \citep{christiano2018supervising}, Recursive Reward Modelling \citep{leike2018scalable}, and market making \citep{hubinger2020marketmaking}. Debate has shown positive results when there is knowledge asymmetry between the AI assistance and the ``judge'' rater, but when this asymmetry is removed, the judge receives little benefit, and adding multiple turns of debate offers no further improvement~\citep{michael2023debate, parrish2022single, parrish2022two, khan2024debating, kenton2024debate}.

Thus far the field has primarily focused on machine learning solutions. Amplified Oversight has not yet considered hybridization, and both Amplified Oversight and HCI have primarily focused on scenarios where AI assistant performance is on par with or surpasses human capabilities.

\subsubsection{Hybridization and Rater Assistance: Interaction Effects}

It is important for Amplified Oversight research to consider Rater Assistance in the context of Hybridization, and the interaction between them. Hybridization might change the kind of assistance we explore---the form of assistance that is most helpful overall on the entire dataset might not be the form that is most helpful on a particular slice. Additionally, for Hybridization, the space of possible signals to ensemble is richer than previously proposed: we can take into account not just AI or human ratings, but also assisted human ratings broken down by different assistant methods. This is a more complicated but also a more flexible and powerful set of signals to optimally combine.

Hybridization has also been studied in the field of Active Learning, which aims to select a subset of data such that training on that subset leads to a model with high accuracy. Amplified Oversight's goal is similar: to improve the quality of training data such that the resulting model is more robust and of higher quality. Common Active Learning strategies include Uncertainty Sampling, where data points with low AI-confidence are selected for human labeling \citep{math11040820}.

\subsection{Additional Details on Evaluation Set, Model, and Experiments}
\label{sec:appendix_model_details}

\subsubsection{Evaluation Set}

\ifchi
\else
The \textit{Evaluation Set} has 1918 [prompt, response, target sentence] tuples in total. The prompts come from a realistic and representative distribution of real user interactions with Gemini, and the responses come from Gemini at the time (late 2023, early 2024), through the Gemini App UI (then called "Bard"). Prompts and responses underwent rigorous filtering of Personally Identifying Information (PII), ensuring researchers only accessed de-identified data. The responses were split into sentences using code based on the NLTK library \cite{bird2009natural}.
\fi

The golden labels come from a high-quality human rating pool specifically trained for fact-verification. Three different raters rate each sentence as either "Accurate" or "Inaccurate". They could also rate sentences as "Doesn't require assessment", however, sentences with at least one of those labels were discarded from our Evaluation Set to ensure that all sentences contained information to fact-verify. We take the majority (i.e., modal) rating out of the 3 labels to get our golden label for each sentence. We have manually reviewed some of these ratings, and we do find them to be very accurate.

Here's a generated representative example of the type of data in the Evaluation Set:

\begin{tcolorbox}[
    title={Representative Eval Set Example},
    colback=blue!5!white, 
    colframe=blue!75!black, 
    fonttitle=\bfseries
]
\textbf{Prompt:} Why are some lakes green? \\
\textbf{AI Response (shortened here, but the full response is shown to raters):} [\dots] Nutrient Enrichment: The primary driver of algal blooms is an abundance of nutrients, particularly phosphorus and nitrogen. These elements act as fertilizers for algae. [\dots] \\
\textbf{Target Sentence:} These elements act as fertilizers for algae. \\
\textbf{Golden Label:} Accurate
\end{tcolorbox}





\subsubsection{The AI Fact-verification model}

\paragraph{Model Description}\label{subsec:model_description}

We built an AI model that uses a search-engine to research the factuality of a given sentence. This model can serve as both a rater and an assistant to human raters. We took inspiration from SOTA AI Fact-verifiers \cite{wei2024long} and trained the AI model to follow these 5 steps:

1. \textbf{Decompose:} The AI decomposes the sentence to be fact-verified into its constituent Factual Claims.

2. \textbf{Search the web:} The AI iteratively searches the web for information that supports and/or that contradicts each factual claim. It uses a Search Tool to issue Search Queries, and for each query it receives a set of (around ~5) Search Results that include the relevant website, and a relevant short snippet from the web page (usually a couple of sentences). It is trained to continue to issue Searches until it finds all necessary information to determine the factuality rating for each factual claim and thus the sentence as a whole.

3. \textbf{Select quotes:} The AI then selects only the most relevant quotes from the snippets as Evidence, along with the corresponding website URL, and numerically lists this evidence. These quotes are verified, verbatim quotes from the websites in the search results.

4. \textbf{Reasoning:} For each decomposed Factual Claim, the AI gives a short explanation for why that claim is factually accurate or not based on the selected evidence and cites those sources. For each claim, it also determines a Factuality Rating.

5. \textbf{Overall Rating:} Finally, the AI determines an Overall Factuality Rating for the sentence based on the ratings for each claim (all Factual Claims need to be accurate for the sentence to be "Accurate").

Claim-level and Overall Ratings could be one of ["Inaccurate", "Unsupported", "Disputed", "Accurate"]. But, the model's Overall Ratings were always one of "Inaccurate" or "Accurate" on our Evaluation Set, perhaps because of how the model was trained.


To determine the final rating of our model for each sentence in our evaluation data-set, we sample our model 50 times, and only select the samples that pass the Format Verifier (described below). On average, there were 33.25 verified samples per sentence. Out of these verified samples, we determine the Model Rating by taking the Majority (modal) Vote of the samples' Overall Ratings. To determine the AI output that we display to human raters in the assistant experiments, we choose the verified sample that has the highest reward model (RM) score (described below), out of those samples that have the same Overall Rating as the Majority Vote Rating (this is called Best-of-N sampling). In our Debate experiment where human raters see two AI outputs one arguing for "Accurate" and one arguing for "Inaccurate", we display the two outputs that each have the highest RM score among the samples with the same rating.

For each sentence, we also determine a model Confidence level. This is done by calculating the proportion of verified samples that have the same Overall Rating as the Majority Vote. Since the model's Overall Rating was binary, this confidence value was between .5 and 1. We found that confidence scores were fairly well-calibrated with accuracy, which is critical for confidence-based hybridization to be effective (See Fig \ref{fig:model_calibration}).

Below is a representative example of the full output of the model after formatting. This formatted output is what is shown to raters. Right-facing triangles indicate HTML <details> tags that can be clicked to expand and reveal further information:

\begin{figure*}[t]
\begin{tcolorbox}[
    title={Representative AI Assistance of the full model output},
    colback=gray!5!white,
    colframe=gray!60!black,
    fonttitle=\bfseries
]
\color{black}
\textbf{Experimental AI-generated fact-verification.} \\
$\triangleright$~\textcolor{orange!80!black}{\faExclamationTriangle} Be careful - this could be misleading
\vspace{2mm} \hrule \vspace{2mm}

\textbf{Factual claims in sentence, and summary of evidence:}\\
\textit{$\triangleright$~(expand to learn more about the information below)}
\vspace{2mm}

\textbf{Claim 1:} Strawberries are a source of Vitamin C. \\
\textbf{Summary of Evidence on claim:} \\
Multiple sources confirm that strawberries are a source of Vitamin C [1, 3]. \\
\textbf{Predicted verdict for claim (could be incorrect):} \textcolor{green!60!black}{\faCheckCircle} \\
Accurate
\vspace{2mm}

\textbf{Claim 2:} The amount of Vitamin C in strawberries is a significant portion of the recommended daily value. \\
\textbf{Summary of Evidence on claim:} \\
Adults need 40 mg/day of Vitamin C [2]. 100g of strawberries provides 58.8mg of Vitamin C [3], so it is considered a good source of the nutrient.\\
\textbf{Predicted verdict for claim (could be incorrect):} \textcolor{green!60!black}{\faCheckCircle} \\
Accurate

\vspace{2mm}
\textbf{Predicted Overall Verdict (could be incorrect):} \textcolor{green!60!black}{\faCheckCircle} Accurate \\
Model Confidence: high (95\%) \\
\textit{$\triangleright$~(expand to understand "confidence")}
\vspace{2mm}

\textbf{Selected Evidence:}\\
\textit{$\triangleright$~(expand to learn more)}
\begin{enumerate}[nosep, leftmargin=*]
    \item \href{https://pmc.ncbi.nlm.nih.gov/articles/PMC4632771/}{\textcolor{blue}{Bioactive Compounds and Antioxidant Activity in Berries - NCBI}}
    \begin{quote}
        Amongst the fruits, fresh strawberries are considered to be one with the highest content of ascorbic acid. Among the berry species, strawberries have similar content to raspberries, but about four-times more ascorbate than blueberries. Ascorbate content in strawberries is highly variable, and in fresh strawberries generally ranges from 5 to 50 mg/100 g fresh weight (fw)
    \end{quote}
    \item \href{https://www.nhs.uk/conditions/vitamins-and-minerals/vitamin-c/}{\textcolor{blue}{Vitamin C - - - Vitamins and minerals}}
    \begin{quote}
        How much vitamin C do I need? Adults aged 19 to 64 need 40mg of vitamin C a day. You should be able to get all the vitamin C you need from your daily diet.
    \end{quote}
    \item \href{https://fdc.nal.usda.gov/fdc-app.html#/food-details/167762/nutrients}{\textcolor{blue}{Strawberries, raw - USDA FoodData Central}}
    \begin{quote}
        Nutrient name: Vitamin C, total ascorbic acid. Amount per 100g: 58.8 mg. Footnotes: Source: USDA Nutrient Data Laboratory. SR Legacy, 2018.
    \end{quote}
\end{enumerate}
\vspace{2mm}

\textbf{All Search Queries and Results:}\\
\textit{(Expand a query to see the results and a quote from each webpage.)}
\begin{itemize}[nosep, leftmargin=*]
    \item $\triangleright$~Search query: \texttt{"strawberries good source of vitamin C"}
    \item $\triangleright$~Search query: \texttt{"recommended daily value of vitamin c"}
\end{itemize}
\end{tcolorbox}
\end{figure*}

\FloatBarrier

\paragraph{Model Training}

The AI assistant fact-verification model was trained by Supervised Fine-Tuning a pre-trained Gemini 1.5 Pro model, using high-quality human-written demonstrations. The human raters writing these demonstrations came from a high-quality rating pool (different from the pool that provided the golden labels), and they followed the same 5 steps as the AI model.

To ensure the human-written demonstrations were of suitable quality, we built a comprehensive tutorial so the human raters would understand the goals and nuances of fact-verification, as well as the steps to follow and the specific format of the model's output. Fact-verification even a single sentence can be a challenging task depending on the nature and number of factual claims in the sentence; there are many edge cases that can be difficult to verify. We provided detailed information and many examples to help raters understand the task and equip them with the skills and knowledge necessary to achieve it.

When writing their demonstrations, the raters had to adhere to the format described above for the AI and use the same search tool as the AI. To ensure each demonstration was formatted correctly, we implemented a Format Verifier to check demonstrations in real-time: each demonstration had to pass this verifier in order for the rater to be able to submit it. The verifier also had the following checks: the explanation for each claim must cite at least one piece of evidence; each piece of selected evidence must be cited somewhere; and the quote per piece of selected evidence must come verbatim from the webpage snippets returned in the search results. 

We also trained a Preference Reward Model (RM). We obtained side-by-side preference ratings from the same human raters who wrote the demonstrations, and used these ratings to train the RM. We used this RM for the  Best-of-N sampling mentioned in Section \ref{subsec:model_description}.

\subsubsection{Human Factuality Rating Task}

We asked the human raters to follow a similar process to our AI rater. For each sentence, raters classified each target sentence as "Inaccurate", "Unsupported", "Disputed", or "Accurate", using online research and/or information from the AI fact-verification assistant (if displayed). For a sentence to be "Accurate", all factual claims in the sentence need to be accurate. A claim is accurate if no reputable contradictory evidence exists; if the claim is an opinion or recommendation, it is accurate if it is reasonable and widely held. If raters believed that a sentence did not contain any factual claims, they could choose a third option of "Doesn't require assessment." According to our golden labels, all sentences contain factual claims, so "Doesn't require assessment" ratings were coded as incorrect labels in our analysis. Lastly, raters could additionally choose to skip sentences they were unable to understand or confidently evaluate (by clicking a Skip button, or rating the sentence as "Can't Confidently Assess"), though they were asked to try to avoid doing so and were given an opportunity to indicate their confidence in the factuality rating (see below).

\subsubsection{Human Experiment Set-Up}

In all experiments, participants first completed a short tutorial that introduced them to the task, explained the study duration (one hour) and payment. We also included a short section explaining why fact-verification AI model outputs is important (i.e., to guard against models generating false, inaccurate, or misleading information that could lead to problematic or unsafe decisions on the part of users). We hoped this explanation would tap into participants' intrinsic motivation and improve their performance, as well as critical engagement with the assistant. Participants were also explicitly encouraged to do their best to provide the correct factuality rating for each sentence and to focus more on quality than quantity (i.e., to spend more time on fewer tasks). There was then a short reading comprehension quiz that provided immediate feedback; participants had to answer all questions correctly to advance. For participants in the unassisted baseline experiments, this was the end of the tutorial and they proceeded to the main task. For participants in one of the assisted intervention experiments, after the quiz, there was an additional short section explaining that they would be shown information from an AI fact-verification assistant. Participants were informed that this fact-verification assistant was still learning and could make mistakes so they should be critical of the information it provided and only use it to the extent they deemed it valid. We hoped this warning would help raters better understand the models' limitations and foster more appropriate reliance.

After completing the tutorial, participants spent the remainder of the hour on the main fact-verification task.
For each fact-verification example, participants saw the AI-generated sentence, and underneath it, the factuality rating scale where they classified the factual accuracy of the sentence as described above. Below the factuality rating scale was a four-point confidence likert scale, where raters could indicate their confidence in the factuality rating as: "not at all confident", "somewhat confident", "mostly confident", "completely confident". This confidence rating was not used in our analysis.

In the assisted intervention experiments, the screen set-up was the same as for the unassisted baseline experiments, except that above the AI-generated sentence and rating scales, information from the AI fact-verification Assistant was displayed in a clearly demarcated, blue box. The box was titled, "Experimental AI-generated Fact-check" with a warning underneath to be careful because the information could be misleading. In addition to rating the factual accuracy of the sentence and their confidence in that rating, participants also rated the helpfulness of the experimental AI-generated fact-verification on a three-point likert scale: "not at all helpful", "somewhat helpful", "extremely helpful". Lastly, there was an optional comments box where participants could share any comments they might have about the AI-generated fact-verification.

Each experiment was between-subjects but within an experiment, participants could take part in the experiment more than once. They only completed the tutorial one time at the start of their first session. Participants were financially compensated for each hour of work spent on the experiment.

\subsection{Statistical Details of Results}

In this section, we detail exactly the statistical tests we use for each result, and their coefficients.

\subsubsection{RQ1 Statistics}
\label{sec:hybrid_results_details}

More details on the \ref{sec:hybrid_results} results: To test the statistical significance of these observed differences, we conducted a mixed effects logistic regression on the entire Evaluation Set predicting label accuracy (1 or 0) from a fixed effect of rating protocol (three-level factor: AI alone, human alone, hybridized, with AI alone as the reference category) with random intercepts by example and by label source (i.e., AI, human). Majority vote human alone ratings were significantly less accurate than AI ratings alone ($\beta = -1.461, SE =  0.161, z = -9.070, p < .001$), but Hybridized accuracy was significantly more accurate than AI ratings ($\beta = 0.413, SE = 0.164, z = 2.520, p = .012$).

\subsubsection{RQ2 Statistics}
\label{sec:rq2_results_details}

More details on the \ref{sec:rq2_results} results: To test if certain differences are statistically significant, we filtered data to the Post-Hybridized Human Set and fit a mixed effects logistic regression predicting label accuracy (1 or 0) from a fixed effect of experiment (10-level factor, with the unassisted baseline\_t2 dummy coded as the reference variable), and a random intercept by participant. This analysis revealed that even though the six experiments manipulating whether the Evidence+Reasoning, Judgements, or Confidence were numerically higher in accuracy than baseline, that these differences were not significant (all $\beta's < 0.168$ and all $p's > 0.225$). For the less leading versions of the assistant, search alone was no different from baseline (search, 68.7\% vs. baseline, 67.3\%; $\beta = 0.072, SE = 0.141, z = 0.511, p = 0.609$), and interestingly, debate was the only intervention that was numerically lower in accuracy than baseline, though this difference was also not significant (64.9\%; $\beta = -0.124, SE = 0.137, z = -0.904, p = 0.3669$). The only form of assistance that led to a significantly higher average accuracy than baseline was search and evidence (73.3\%, $\beta = 0.308, SE = 0.136, z = 2.269, p = 0.023$).

If we re-run this analysis with search and evidence dummy coded as the reference variable we find that Search and Evidence performs significantly better than Judgments \& Confidence ($\beta = -0.285, SE = 0.141, z = -2.024, p = 0.043$) and Debate ($\beta = -0.432, SE = 0.145, z = -2.981, p = 0.003$) but no different than the other interventions (all $\beta$'s between $-0.236$ and $-0.141$ and all $p's > 0.111$).

\subsubsection{RQ2a Statistics}
\label{sec:rq2a_details}

More details on the \ref{fig:overtrust_interventions} over-reliance results: To test whether raters over-relied on particular forms of assistance, we fit a mixed effects logistic regression predicting label accuracy (1 or 0) from fixed effects of experiment (10 level factor, with the unassisted baseline dummy coded as the reference variable) and model accuracy (two level factor, with incorrect as reference), and their interaction with a random intercept by participant.

When the model was incorrect, the decrement in performance was statistically significant for the following conditions:
\begin{itemize}
    \item \textbf{Evidence \& Reasoning \& Judgments \& Confidence:} $\beta = -0.636, SE = 0.187, z = -3.405, p < .001$
    \item \textbf{Evidence \& Reasoning \& Judgments:} $\beta = -0.768, SE = 0.174, z = -4.413, p < .001$
    \item \textbf{Judgments \& Confidence:} $\beta =  -0.356, SE = 0.176, z = -2.029, p = .042$
\end{itemize}

For conditions where just the overall Judgment was shown alone, or if no Judgments were included, assisted performance was numerically but not statistically worse than baseline (all $\beta$'s between $-0.343$ and $-0.172$ and all $p's > .054$).

\subsection{Supplementary Results}


\subsubsection{Terminology}

The Supplementary Results below use a different, older terminology: "Trace" refers to the combination of Search Results and Evidence mentioned in Section \ref{sec:t2}. "Ratings" refers to Judgements. "Confidence" means the same. "+" means one of those is included in the assistance, and "-" means that it is not included.

\subsubsection{How do monetary incentives change how the assistance helps?}
\label{sec:bonus_desc}

\begin{figure*}[h]
  \centering
  \includegraphics[width=0.7\textwidth]{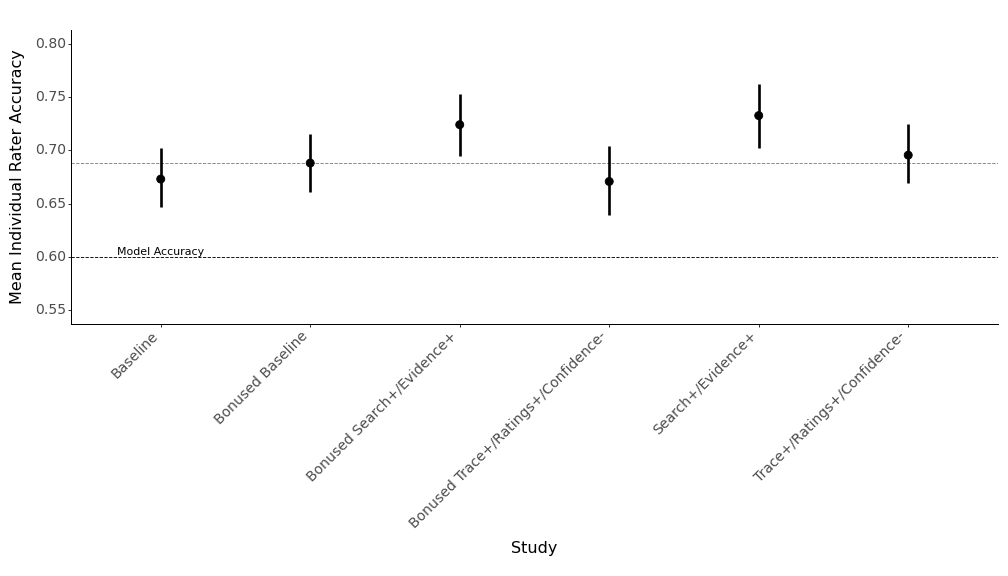}
  \caption{Mean Individual Rater Accuracy for the bonus and select T2 studies described in Sections \ref{sec:bonus_desc} and \ref{sec:t2}. Restricted to examples where model confidence <= .62.}
  \label{fig:bonus_indiv_acc}
  \Description{A point and error bar chart showing the "Mean Individual Rater Accuracy" for several experimental conditions, comparing those with a monetary bonus to those without. The y-axis ranges from 0.55 to 0.80, and the x-axis lists six conditions (Baseline, Bonused Baseline, Search+/Evidence+, etc.). A dotted line indicates the model's accuracy at 60\%. The error bars for bonused and non-bonused versions of the same condition heavily overlap, indicating that providing a bonus for accuracy did not have a statistically significant effect on overall performance.}
\end{figure*}

In addition to comparing performance using different types of assistance, we also investigated how giving raters a bonus for providing correct ratings impacted the effects of assistance. Prior research has shown that increasing the monetary incentive for performance reduces over-reliance \cite{vasconcelos2023explanations}. To test the effect of a bonus on human rater accuracy, we replicated the unassisted baseline experiment, as well as AI-assistance with Evidence, Reasoning, and Judgments (called \textbf{Trace+}/\textbf{Ratings+}/\textbf{Confidence-} below) and AI-assistance with Search and Evidence only (called \textbf{Search+}/\textbf{Evidence+} below). If participants completed at least six ratings (that were not "Doesn't require assessment") and achieved at least 80\% accuracy on those ratings, they would receive a bonus equal to their hourly rate. The Baseline bonus study was done at the same time as the T2 studies described in Section \ref{sec:t2}. The assisted studies with bonus were done a week later. Each of the T2 and bonus studies were all done with mutually exclusive sets of raters. The resulting Mean Individual Accuracy for these studies are shown in Fig \ref{fig:bonus_indiv_acc}.

Providing raters with a bonus for accuracy did not increase their overall accuracy on the Post-hybridized Human Set with or without assistance. For the baseline experiments, non-bonus accuracy was 67.3\%, while bonus accuracy was numerically but not statistically higher, 68.9\% ($\beta = -0.132, SE = 0.128, z = -1.033, p = .302$). For both \textbf{Trace+}/\textbf{Ratings+}/\textbf{Confidence-} and \textbf{Search+}/\textbf{Evidence+} assistance, bonus accuracy was actually numerically \textit{lower} than non-bonus accuracy, but again these differences were not statistically significant (\textbf{Trace+}/\textbf{Ratings+}/\textbf{Confidence-}: 67.1\% with bonus vs. 69.8\% without bonus, $\beta = 0.084, SE = 0.137, z = 0.612, p = .541$; \textbf{Search+}/\textbf{Evidence+}: 72.4\% with bonus vs. 73.3\% without bonus, $\beta = 0.047, SE = 0.145, z = 0.320, p = .749$). In addition, though \textbf{Search+}/\textbf{Evidence+} assistance increased overall accuracy compared to unassisted baseline without bonus, this assistance with a bonus did not statistically significantly improve accuracy compared to baseline with a bonus (72.4\% vs. 68.9\%, $\beta = 0.130, SE = 0.138, z = 0.939, p = .348$).

\begin{figure*}[h]
  \centering
  \includegraphics[width=\linewidth]{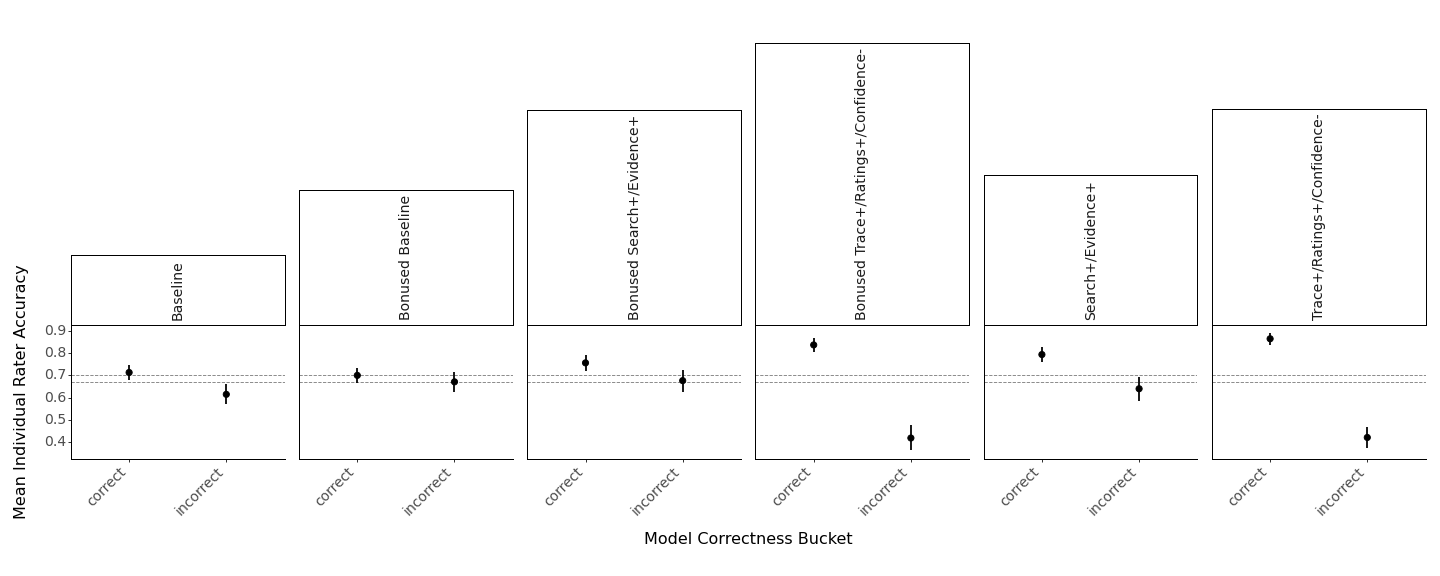}
  \caption{Mean Individual Rater Accuracy for the bonus and select T2 studies described in Sections \ref{sec:bonus_desc} and \ref{sec:t2}, split by if the fact-verification assistant was correct. Restricted to examples where model confidence <= .62.}
  \label{fig:overtrust_interventions_bonus}
  \Description{A series of six small point-and-error-bar plots, arranged horizontally. Each plot corresponds to an experimental condition from Figure 5 and shows "Mean Individual Rater Accuracy" on the y-axis, split by whether the AI assistant was "correct" or "incorrect" on the x-axis. The chart shows that for the baseline condition, a monetary bonus slightly increased accuracy on examples where the model was incorrect. For assisted conditions, the bonus did not reduce over-reliance; for the "Trace+/Ratings+/Confidence" condition, accuracy on incorrect examples was significantly worse with a bonus.}
\end{figure*}

Fig \ref{fig:overtrust_interventions_bonus} shows how under- and over-reliance changes when incentivizing raters for accuracy. Comparing baseline with and without bonus, we find that accuracy when the model is incorrect is higher with a bonus (67.1\% vs. 61.4\%, $\beta = -0.342, SE = 0.174, z = -1.969, p = .049$), so even though overall accuracy was not improved by incentivizing raters for accuracy, this extra monetary incentive did appear to improve performance on harder examples (unassisted performance is lower for model incorrect vs. correct examples suggesting that the examples the model got wrong were also more challenging for humans). However, unlike prior work, incentivizing raters did not appear to reduce over-reliance. In particular, \textbf{Trace+}/\textbf{Ratings+}/\textbf{Confidence-} assisted performance on incorrect examples was still statistically significantly worse when raters were bonused ($\beta = -1.108, SE = 0.184, z = -6.006, p < .001$). Likewise, the difference in accuracy between correct and incorrect examples was no different for \textbf{Trace+}/\textbf{Ratings+}/\textbf{Confidence-} with and without assistance ($\beta = 0.182, SE = 0.237, z = 0.770, p = .442$). For \textbf{Search+}/\textbf{Evidence+} assistance, there was again no evidence of over-reliance (i.e., unassisted and bonused performance was no different from assisted and bonused performance on the incorrect examples, $\beta = -0.060, SE = 0.188, z = -0.318, p = .750$).

In sum, incentivizing raters for accuracy decreased the positive effect of assistance but did not significantly increase overall accuracy nor reduce over-reliance. These results raise questions of how assistance might impact higher quality raters.

\subsubsection{How do the different forms of assistance affect the time spent per task?}

\begin{figure*}[h]
  \centering
  \includegraphics[width=0.7\textwidth]{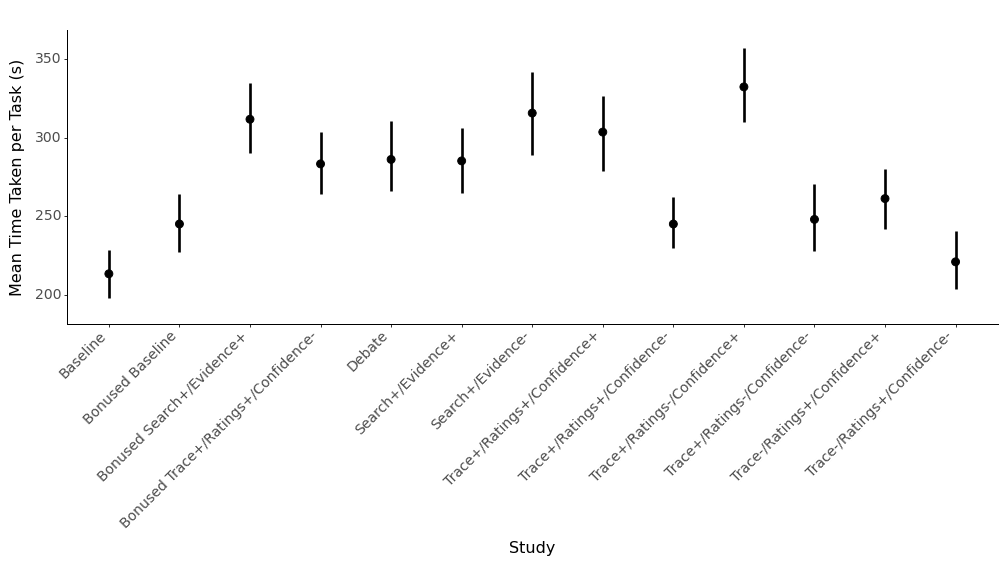}
  \caption{Mean Time Taken per Task. Restricted to examples where model confidence <= .62. These were from "T2" studies, described in Section \ref{sec:t2}. In this plot and below, "Trace" refers to "Search+Evidence+Reasoning".}
  \label{fig:all_duration}
  \Description{A point and error bar chart showing the "Mean Time Taken per Task (s)" on the y-axis, for various experimental conditions on the x-axis. The y-axis ranges from 200 to 350 seconds. The baseline conditions show the shortest times. All forms of AI assistance increase the time taken per task, with the "Search+/Evidence-" condition (showing only search results) taking the longest.}
\end{figure*}

Fig \ref{fig:all_duration} shows that all interventions increase the time raters take to do each task. We did filter out a few outliers in the data of task times that took longer than 1 hour.

There do seem to be certain presentation styles that increase the time taken more than others. Providing just Search results take the longest, possibly because that section has the most content and raters are reading those carefully when only they are presented. Showing evidence also leads to much higher time taken per task compared to baseline and other some interventions, perhaps because raters are going more often to the linked sites and understanding the evidence. There also seems to be a trend of how showing the Confidence increases the amount of time taken. This effect seems stronger when the Trace is present.

The bonus also seems to slow down raters, about the same absolute amount for each presentation style.

\subsubsection{RQ3: Does the optimal confidence threshold for hybridization change with assistance, and if so, how does that affect which form of assistance is best?}
\label{sec:RQ3}

\begin{figure*}[h]
  \centering
    \includegraphics[width=0.9\textwidth]{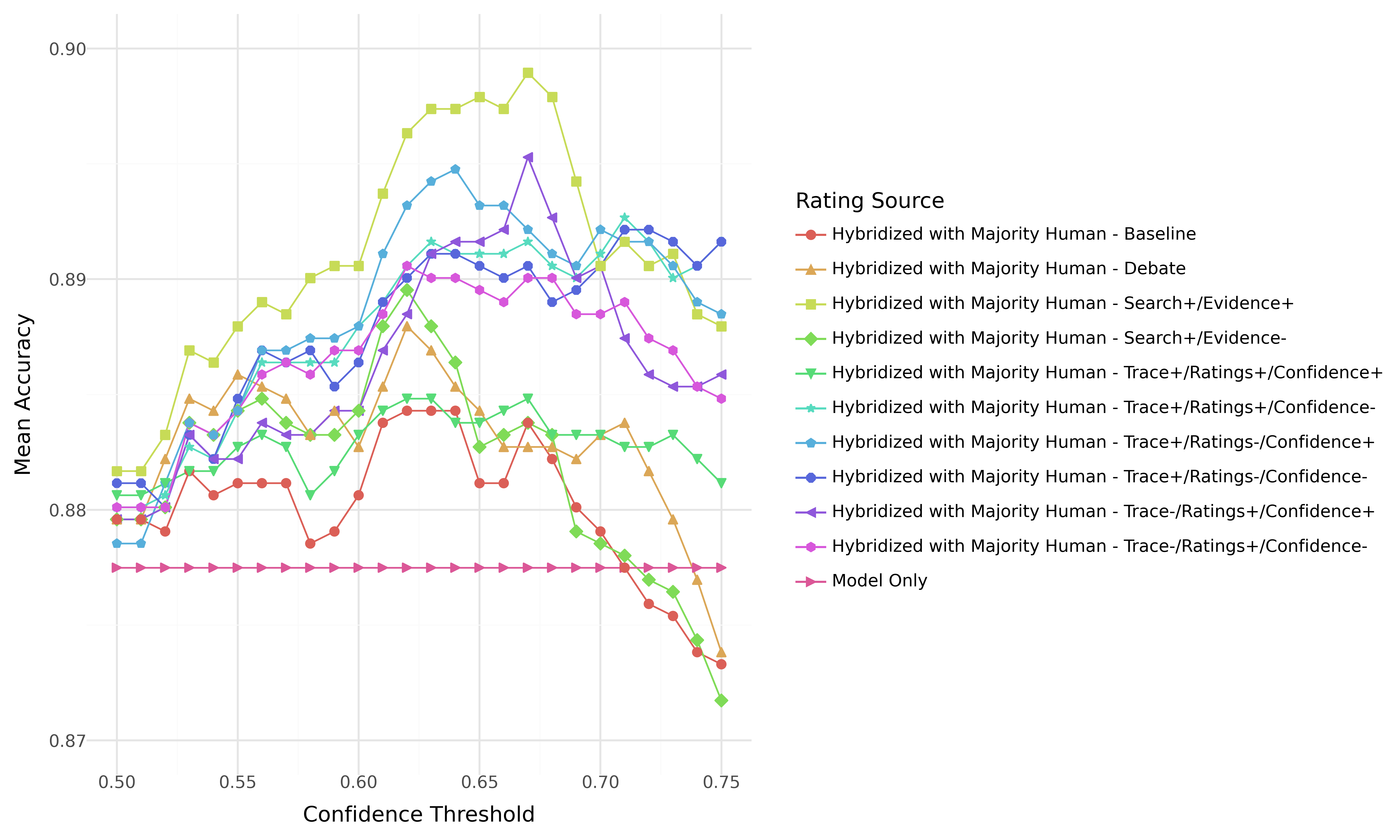}
      \caption{Mean Rater Accuracy, using hybridization with Individual Human ratings for T2 studies (described in Section \ref{sec:t2}).}
  \label{fig:t2_individual_conf_threshold_hybrid}
  \Description{A line chart titled "Mean Rater Accuracy, using hybridization with Individual Human ratings for T2 studies." The x-axis is "Confidence Threshold" from 0.50 to 0.75, and the y-axis is "Mean Accuracy" from 0.87 to 0.90. The chart displays numerous noisy, overlapping lines, each representing a different type of AI assistance combined with human ratings. The chart is exploratory and suggests that different forms of assistance might have different optimal hybridization thresholds, motivating further experiments.}
\end{figure*}

In our initial experiment for RQ1, we fixed the kind of human ratings (unassisted baseline), and we allowed the confidence threshold to vary to identify the optimal threshold for hybridization. In the experiments for RQ2, we fixed the confidence threshold at the previously selected optimal level (T=.62) and explored various types of human ratings, unassisted and different forms of assisted. Now, we run a series of experiments that allow us to vary both the confidence threshold and the human ratings. Varying both the threshold and type of assistance will enable us to see if the optimal threshold is different for different forms of assistance and whether this changes the optimal assistance. Figure \ref{fig:t2_individual_conf_threshold_hybrid} generated with T2 results shows how changing the threshold might increase accuracy for certain types of assistance. These results are limited in that we cannot increase the threshold beyond .75, and we don't have enough ratings per example to see benefit from majority vote rating, leading us to run another study.

We select three forms of assistance to explore: (1) trace with ratings and confidence (\textbf{Trace+}/\textbf{Ratings+}/\textbf{Confidence+}), (2) trace with ratings but no confidence (\textbf{Trace+}/\textbf{Ratings+}/Confidence-), and (3) search and evidence (\textbf{Search+}/\textbf{Evidence+}). We select search and evidence because it is our most promising form of assistance (at least with T=.62). We select the other two because they helped the most when the overall rating was correct (though they also hurt the most when this rating was incorrect). It’s thus possible that these forms might be more helpful at higher thresholds where a higher proportion of model overall ratings are correct. We can also explore whether hybridizing with multiple forms of assistance (e.g., search and evidence for low confidence, trace with ratings and confidence for medium confidence, and AI alone for high confidence) achieves an even higher accuracy than hybridizing with a single form of assistance.



\begin{figure*}[h]
  \centering
  \begin{minipage}[b]{0.49\linewidth} 
    \centering
    \includegraphics[width=\linewidth]{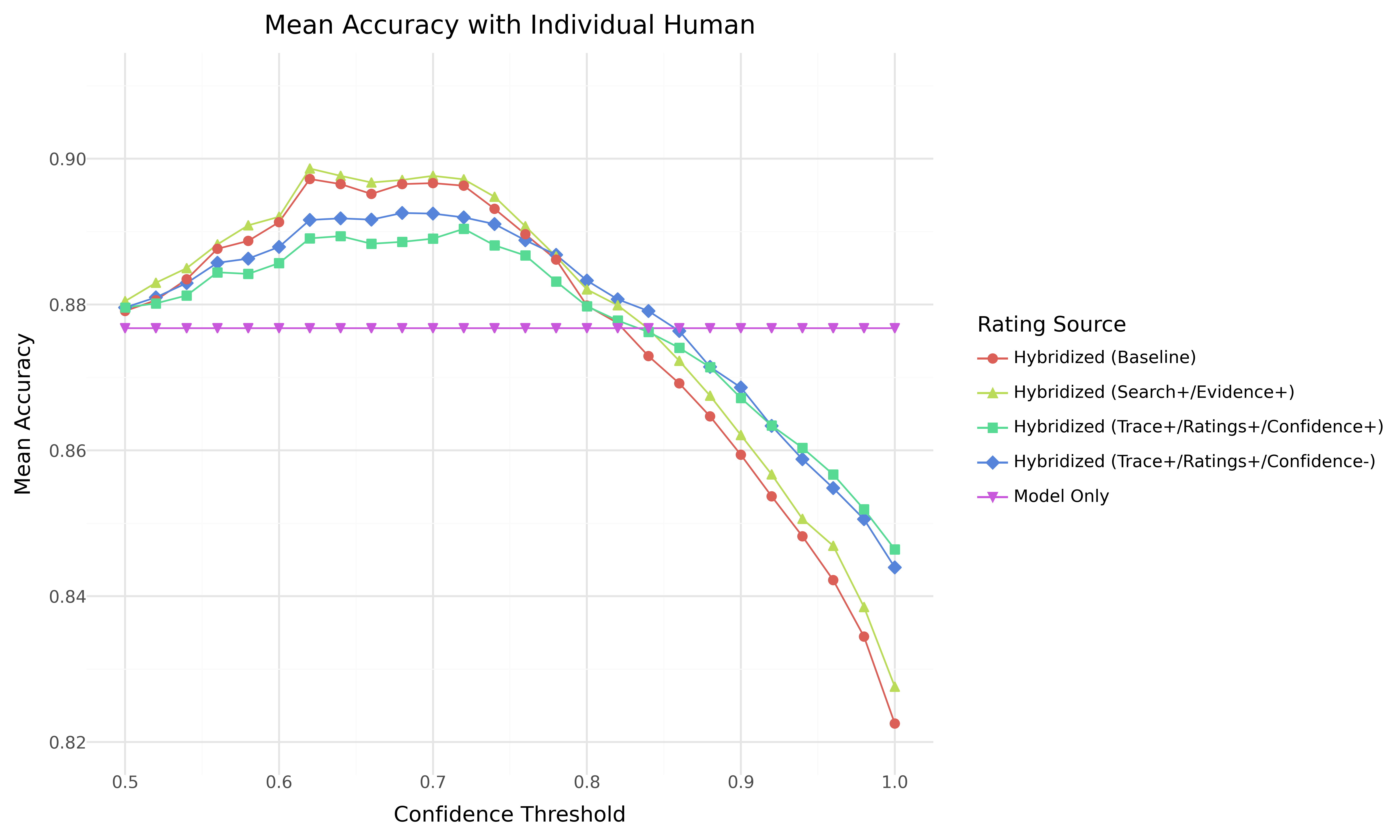}
      \caption{Mean Rater Accuracy, using hybridization with Individual Human ratings, for "T3" studies, described in Section \ref{sec:t3}.}
  \label{fig:all_test_individual_conf_threshold}
  \Description{A line chart titled "Mean Accuracy with Individual Human." The x-axis is "Confidence Threshold" and the y-axis is "Mean Accuracy." It shows the performance of a "Model Only" approach compared to four different hybridized approaches using individual human ratings. The hybridized approaches all achieve higher accuracy than the model alone at confidence thresholds below approximately 0.8. Their peak performance is clustered between a threshold of 0.6 and 0.7, after which their accuracy declines.}
  \end{minipage}
  \hfill 
  \begin{minipage}[b]{0.49\linewidth} 
    \centering
    \includegraphics[width=\linewidth]{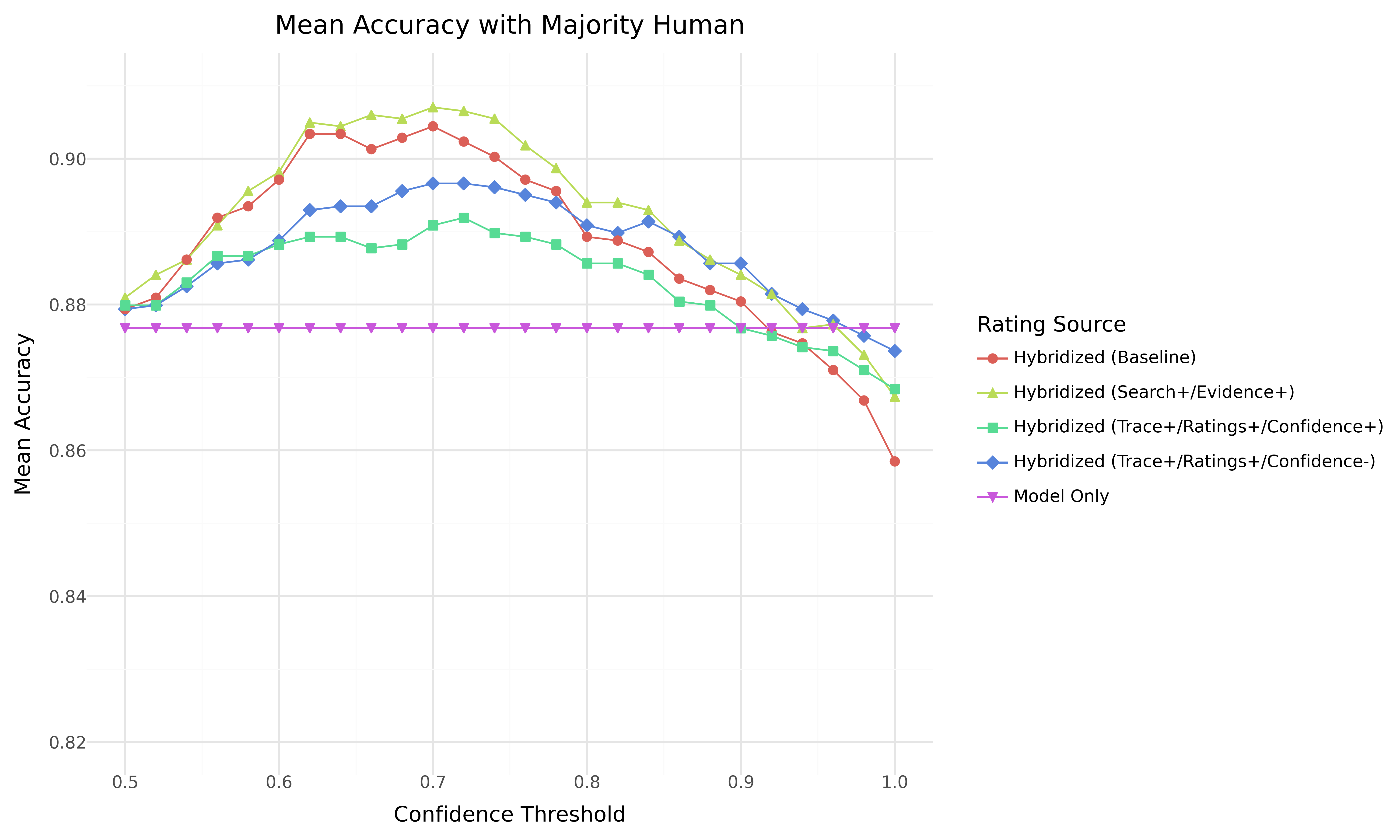}
      \caption{Mean Rater Accuracy, using hybridization with Majority Human vote, for "T3" studies, described in Section \ref{sec:t3}.}
    \label{fig:all_test_majority_conf_threshold} 
    \Description{A line chart titled "Mean Accuracy with Majority Human." The chart is structurally identical to Figure 8 but uses majority vote human ratings, resulting in higher overall accuracy on the y-axis. The "Hybridized (Search+/Evidence+)" and "Hybridized (Baseline)" lines achieve the highest accuracy, peaking at over 90\% near a confidence threshold of 0.7. This shows that aggregating human ratings improves the performance of the hybridized system.}
  \end{minipage}
\end{figure*}

\paragraph{Human (Rater) Experiments}
\label{sec:t3}

The experiments were identical in structure and content to the prior \textbf{Search+}/\textbf{Evidence+}, \textbf{Trace+}/\textbf{Ratings+}/\textbf{Confidence+}, and \textbf{Trace+}/\textbf{Ratings+}/Confidence- experiments except that all together participants rated the entire Evaluation Set (1918 examples). We also made the experiments available to the entire pool of participants to ensure that we could get a sufficient number of ratings per example on the entire set. This means that participants could take part in multiple experiments. We also ran a fresh baseline experiment (baseline\_t3 for time 3) to match the conditions under which the intervention experiments were being launched, and also to account for performance increases due to prior experience with our task, as well as other teams' factuality tasks. We recruited for 800 rating hours per experiment, and participants had an opportunity to take part two times, spots permitting. Average number of ratings per example was 3.61 for each experiment (we clamped the maximum number of ratings per example to be the minimum across all "T3" experiments).


\paragraph{Results}

It appears that the human raters unassisted baseline performance did improve over time. In the new baseline experiment (baseline\_t3), human raters achieved 82.7\% accuracy on the entire Evaluation Set based on individual labels and 85.8\% accuracy based on majority labels, whereas in the initial baseline experiment (baseline\_t1) accuracy was 75.1\% and 84.7\%, respectively. Additionally, if we filter this data to the Post-Hybridized Human Set (T = .62), we see that baseline\_t3 accuracy (74.1\%) is also higher than baseline\_t2 accuracy (67.3\%). The baseline\_t3 performance was still lower than the AI rater alone, which, as a reminder, achieves 87.7\% accuracy on the Evaluation Set. However, it's important to keep in mind that the RQ3 experiments involve higher performing raters than RQ1 and RQ2.

In the current experiments, we again find evidence that confidence-based hybridization improves accuracy above human ratings or AI ratings alone, achieving complementary performance. Confidence-based hybridization with T=.62 and using unassisted (baseline\_t3) ratings achieves 89.7\% (individual) and 90.3\% accuracy (majority). However, confidence-based hybridization with assistance did not lead to an additional improvement, likely because baseline performance was quite high. In fact, \textbf{Trace+}/\textbf{Ratings+}/\textbf{Confidence+} and \textbf{Trace+}/\textbf{Ratings+}/Confidence- assistance slightly decreased overall hybridized accuracy, achieving 88.9\% (majority) and 89.3\% (majority), respectively. \textbf{Search+}/\textbf{Evidence+} assistance did not make a difference, achieving 90.5\% (majority).

Indeed, we find that on the Post-Hybridized Human Set, unlike RQ2, \textbf{Search+}/\textbf{Evidence+} does not increase accuracy above baseline and in some cases, \textbf{Trace+}/\textbf{Ratings+}/\textbf{Confidence+} and \textbf{Trace+}/\textbf{Ratings+}/Confidence- assistance actually leads to a statistically significantly worse accuracy than baseline. To test these relationships, we filter the data to the T=.62 Post-Hybridized Human Set and fit a mixed effects logistic regression predicting individual rating accuracy (0 or 1) from a fixed effect of label-type (5-level categorical variable: AI, unassisted baseline\_t3 human, \textbf{Search+}/\textbf{Evidence+} assisted human, \textbf{Trace+}/\textbf{Ratings+}/\textbf{Confidence+} assisted human, and \textbf{Trace+}/\textbf{Ratings+}/Confidence- human, with unassisted baseline\_t3 as the reference) and a random intercept by conversation. This analysis indicates that while \textbf{Search+}/\textbf{Evidence+} and \textbf{Trace+}/\textbf{Ratings+}/Confidence- are no different than baseline ($\beta$'s between -0.208 and 0.112, $p's >$.06), \textbf{Trace+}/\textbf{Ratings+}/\textbf{Confidence+} is statistically significantly worse than baseline (74.1\% vs. 70.0\%; $\beta = -0.317, SE = 0.110, z = -2.876, p = .004$). Note, that all humans ratings (unassisted or assisted) are statistically significantly better than the AI ratings, which is again why we find that hybridization works.

But what happens if we explore different thresholds above T=.62? Figures \ref{fig:all_test_individual_conf_threshold} and \ref{fig:all_test_majority_conf_threshold} show how mean hybridized accuracy changes when using baseline and each form of assistance across different confidence thresholds. We find that (1) the optimal threshold does not vary much across the different forms of assistance, (2) it is never greater than .75, and (3) the optimal form of assistance does not change. For \textbf{Search+}/\textbf{Evidence+} assistance the optimal thresholds are the same as in RQ2, suggesting that these were not local optima: T=.62 for individual ratings (89.9\%) and T=.7 (90.7\%) for majority ratings. For \textbf{Trace+}/\textbf{Ratings+}/Confidence-, the optimal threshold is T=.68 for individual ratings (89.3\%) and T=.7 for majority ratings (89.7\%). For \textbf{Trace+}/\textbf{Ratings+}/\textbf{Confidence+}, the optimal threshold is T=.72 both for individual (89.0\%) and majority (89.2\%). Moreover, we do not find a case where assisted overall accuracy is appreciably better than baseline: We achieve the highest accuracy with T=.7 hybridizing using either majority vote baseline (90.4\%) \textit{or} majority vote \textbf{Search+}/\textbf{Evidence+}) assistance (90.7\%).

It appears that with more skilled raters, the benefits of even our most promising assistant (\textbf{Search+}/\textbf{Evidence+}) are reduced. If we filter data to the optimal post-hybridized set for this form of assistance (T=.7), which is also optimal for \textbf{Trace+}/\textbf{Ratings+}/\textbf{Confidence+}, we find that \textbf{Search+}/\textbf{Evidence+} does not statistically significantly improve accuracy above baseline ($\beta = 0.084, SE = 0.183, z = 0.457, p = .647$). \textbf{Trace+}/\textbf{Ratings+}/Confidence- also has no effect ($\beta = -0.239, SE = 0.179, z = -1.336, p = .182$), while \textbf{Trace+}/\textbf{Ratings+}/\textbf{Confidence+} statistically significantly decreases accuracy below baseline majority-vote ratings on this set ($\beta = -0.404, SE = 0.177, z = -2.281, p = .023$).

In sum, we again find evidence that confidence-based hybridization improves accuracy above and beyond using AI ratings. Hybridizing with unassisted majority vote, or \textbf{Search+}/\textbf{Evidence+} assisted majority vote human ratings yields a 3\% increase above using AI ratings alone. Unlike our RQ2 experiments, we do not find evidence that assistance further increases accuracy on top of the gains from hybridization, and interestingly, the most naive form of assistance that just shows human raters everything that the AI fact-verification assistant outputs (\textbf{Trace+}/\textbf{Ratings+}/\textbf{Confidence+}), hurts performance. These findings suggest that the effectiveness of assistance may vary depending on rater skill, and that assistance tested on one type of rater may not transfer to other raters or even the same raters at a later time. 

\FloatBarrier



\end{document}